\documentclass{article}

\usepackage[preprint]{corl_2021} 

\usepackage{subcaption,graphicx}
\usepackage{overpic}
\usepackage{wrapfig}
\usepackage{float}
\usepackage[font=small]{caption} 
\usepackage{tabularx}
\usepackage{paralist}
\newcommand{\tabitem}{~\llap{\textbullet}~}
\usepackage{multirow}
\usepackage{makecell}
\newcommand{\STAB}[1]{\begin{tabular}{@{}c@{}}#1\end{tabular}}

\usepackage{outlines}

\usepackage{siunitx}

\usepackage[numbers]{natbib}
\usepackage{multicol}

\title{Tactile Sim-to-Real Policy Transfer via Real-to-Sim Image Translation}

\author{
  Alex Church\\
  Department  of  Engineering  Mathematics\\
  Bristol  Robotics  Laboratory\\
  University  of  Bristol,
  U.K.\\
  \texttt{ac14293@bristol.ac.uk} \\
\And
  John Lloyd\\
  Department  of  Engineering  Mathematics\\
  Bristol  Robotics  Laboratory\\
  University  of  Bristol,
  U.K.\\
  \texttt{jl15313@bristol.ac.uk} \\
\AND
  Raia Hadsell\\
  Google Deepmind\\
  U.K.\\
  \texttt{raia@google.com} \\
\And
  Nathan F. Lepora\\
  Department  of  Engineering  Mathematics\\
  Bristol  Robotics  Laboratory\\
  University  of  Bristol,
  U.K.\\
  \texttt{n.lepora@bristol.ac.uk} \\
}

\begin{document}
\maketitle
\vspace{-2em}

\begin{abstract}
    Simulation has recently become key for deep reinforcement learning to safely and efficiently acquire general and complex control policies from visual and proprioceptive inputs. Tactile information is not usually considered despite its direct relation to environment interaction. In this work, we present a suite of simulated environments tailored towards tactile robotics and reinforcement learning. A simple and fast method of simulating optical tactile sensors is provided, where high-resolution contact geometry is represented as depth images. Proximal Policy Optimisation (PPO) is used to learn successful policies across all considered tasks. A data-driven approach enables translation of the current state of a real tactile sensor to corresponding simulated depth images. This policy is implemented within a real-time control loop on a physical robot to demonstrate zero-shot sim-to-real policy transfer on several physically-interactive tasks requiring a sense of touch. 
    
    \small{Video results:
    \href{https://sites.google.com/my.bristol.ac.uk/tactile-gym-sim2real/home}{\emph{https://sites.google.com/my.bristol.ac.uk/tactile-gym-sim2real/home}}. \newline Code: \href{https://github.com/ac-93/tactile_gym}{\emph{https://github.com/ac-93/tactile\_gym}}.
    }
\end{abstract}

    \begin{figure}[!htb]
      \centering
      \scriptsize
      \begin{overpic}[width=1.0\linewidth]{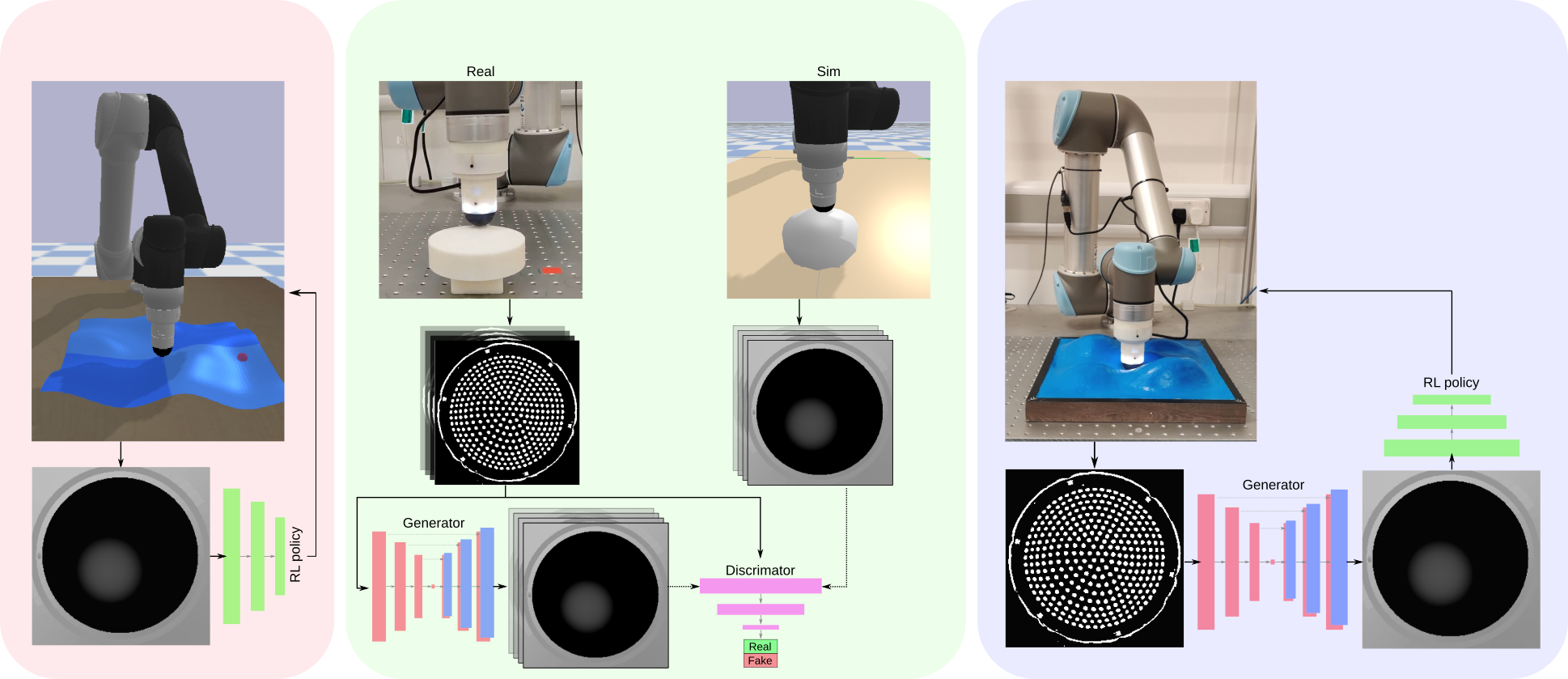}
      \put(1, 40){\parbox{25mm}{\centering (a) Reinforcement Learning in Sim}}      \put(22, 40){\parbox{55mm}{\centering (b) Dataset Collection and GAN Training}}
      \put(61, 40){\parbox{55mm}{\centering (c) Real Robot Evaluation}}
      \end{overpic}
      \caption{Overview of the proposed approach for sim-to-real transfer of learned tactile policies. a) Learn a policy in simulation directly from simulated tactile images. b) Train a GAN for translating between real and sim images using a dataset of image pairs collected by sampling a static environment with features similar to the environment used for reinforcement learning. c) Evaluate on the real robot by passing real images through the generator and then generated images through the RL policy.}
      \label{fig:overview_full}
      \vspace{0.5em}
    \end{figure}

\section{Introduction} \label{sec:introduction}

Learning algorithms have an innate appeal for robotics to enable general, complex behaviours that would be difficult to achieve with classic control methods. Large-scale data is often required for robot learning, and so physics-based simulation has a vital role to play for data collection. A common approach in learning-based robotics is to simulate data that would be impractical to collect in reality, learn control policies from this data, and then transfer learned skills to the physical system. Simulation also offers advantages such as avoiding damage during exploratory training, exploiting privileged information from the simulator, and use of open-sourcing for dissemination of research findings. However, physics engines necessarily approximate the real world to reduce computational costs, giving a `sim-to-real gap' that impairs the performance of these policies applied to reality.

Reinforcement Learning (RL) robotics research is dominated by the use of proprioception and vision as the sensory inputs. Whilst this can lead to complex behaviours \cite{akkaya2019solving, Leeeabc5986, IbarzRobotTrain}, other important sources of information have been underutilised. Specifically, humans use the sense of touch to accomplish complex manual tasks, utilising a granularity of detail unavailable to our other senses. Tactile data has several advantages as the main source of information for learning: it does not suffer from occlusion, particularly for fine manipulation; contact information can be more detailed than visual images of an entire scene; and the observation space is constrained (e.g. to tactile images of markers), simplifying the translation of real to simulated data and, likewise, from simulated to real policies.

Here we make the following contributions aimed at bringing together reinforcement learning and tactile robotics: \textbf{1)} We provide an open-source suite of RL environments tailored to tactile robotics, utilising a simple and fast method of simulating a tactile sensor. \textbf{2)} We show that an RL method, Proximal Policy Optimisation (PPO) \cite{Schulman2017ProximalAlgorithms}, acquires successful policies for all of these tasks.  \textbf{3) } We demonstrate and validate a data-driven approach for zero-shot sim-to-real policy transfer via image translation between real and simulated tactile images on these tasks.

\vspace{-.5em}
\section{Related Work} \label{sec:related_work}

\textbf{Sim-to-Real Transfer:} For computational efficiency, physics engines necessarily approximate the real-world dynamics, leading to a sim-to-real gap where dynamics and visuals differ between simulation and reality. This gap can make it difficult to transfer skills learned in simulation to the real world. Several methods have been proposed to mitigate this issue, namely \emph{Domain Randomisation}, \emph{Network Distillation} and \emph{Domain Adaption} \cite{IbarzRobotTrain}. 

In this work we mainly use \emph{Domain Adaption}, following \citet{james2019sim} to train an image-conditioned generator network that translates between real and simulated images. When using vision, \emph{Domain Randomisation} was necessary on simulated images to train a generator robust enough to generalise to real visual images. Here we exploit the more constrained image space of optical tactile sensors to train generator networks using a controlled tactile dataset. We perform some randomisations to make simulated tasks more difficult and RL policies more robust, aiming to keep within realistic conditions. Although privileged information from the simulator is used to construct a reward for improved learning, this information is not part of the observation in any task. As we perform zero-shot policy transfer from sim-to-real, \emph{Network Distillation} is not required.

\textbf{Tactile Sim-to-Real:} Recently, several works have aimed to bridge between tactile simulation and reality. These split into two categories: simulating the physics of sensor deformation via Finite Element (FE) methods or replicating the captured sensor information via image rendering techniques. A concise overview of the sim-to-real approach applied to non-vision based sensors is given in \cite{narang2021sim}. 

FE methods have seen recent advancements in computational efficiency and accuracy. \citet{narang2021interpreting} simulated deformations of the BioTac sensor that, in combination with data-driven learning, can accurately regress the sensor output. Later, a 75-fold improvement in simulation speed was achieved with GPU-acceleration \cite{narang2021sim}, with each simulated sensor output taking up to 5 seconds which is still impractical for sample-inefficient methods such as DRL. A similar approach was applied to a marker-based optical tactile sensor  for regressing contact force fields \cite{sferrazza2020learning, sferrazza2020sim}. Later work \cite{bi2021zero} achieved zero-shot sim-to-real transfer of a complex policy learned in simulation via RL; however, task-specific approximation was needed for computational efficiency, e.g. assuming planar motion.

Most comparable to our work, \citet{Ding2020Sim-to-RealSensing} use an elastic deformation approach on the same tactile sensor as used here (a TacTip soft biomimetic optical tactile sensor~\cite{Ward-Cherrier2018,lepora2021}). They focused on supervised learning from simulated marker positions rather than tactile images, and with domain randomisation could accurately predict edge position and orientation for use in contour following. The authors noted that the simulation speed is restricted by Unity’s collision detection module.

A computationally efficient approach is to render images from specific optical tactile sensors with rigid-body physics approximating the contact dynamics. Depth images from a GelSight tactile sensor were rendered using the Gazebo physics engine with additional Gaussian smoothing, light rendering and image calibration to help match real tactile images \cite{gomes2021generation}. On a supervised classification task, a drop of 39\% accuracy was found for direct sim-to-real transfer, reduced to 6.5\% with texture augmentation as a form of domain randomisation. A similar approach is proposed in \cite{Wang2020TACTO} with extended image-rendering techniques applicable to a range of GelSight-type sensors on more complex physically-interactive tasks; however, so far no sim-to-real quantitative results have been reported. 

Our work here focuses on a high-resolution tactile simulation environment appropriate for RL of complex control policies. By relying on rigid-body approximations to the contact dynamics, our physics simulation is significantly faster than FE methods without task-specific approximations. Our simulated TacTip images use depth image-rendering techniques like those used for the GelSight \cite{gomes2021generation, Wang2020TACTO}. However, a key novelty is that our simulated depth images do not closely resemble the real TacTip images, because depth is non-trivially related to marker shear for the TacTip (and closely related to image shading for the GelSight). Instead, we close the sim-to-real gap by relying on a later domain adaption phase. Hence, we keep the simulation agnostic to the sensor transduction, which should facilitate future application of these methods to other types of tactile sensor.
 
\vspace{-.5em}
\section{Tactile Simulation} \label{sec:tactile_sim}

In this work, we utilise PyBullet's synthetic camera rendering to capture depth images within a virtual optical tactile sensor, based on the CAD files used to 3D print a real TacTip (see e.g. \cite[Fig. 4]{Ward-Cherrier2018}). When gathering tactile images, we take the difference between the current depth image and a reference depth image taken from when the sensor is not in contact. This difference produces a penetration depth map that generalises to arbitrary sensor shapes. Noise is removed from the image by zeroing values below a set tolerance of $10^{-4}$, followed by a re-scaling from $[0, \mathrm{max\_penetration}]$ to $[0,255]$. An artificial border is also overlaid onto the tactile image to bring it closer to real tactile images and to provide a reference point that transforms with augmentations.

In order to achieve the computational efficiency necessary to generate data at large scales, we approximate the soft tip of the tactile sensor with rigid body physics simulation. We limit the contact stiffness and damping used during collision detection as this allows penetration of objects into the simulated sensor tip in an approximate manner to the deformation of the real tip.

We describe our simulation method in relation to the list of desiderata proposed by 
\citet{Wang2020TACTO}.

\textbf{High Throughput:} We use PyBullet's GPU rendering functionality to offer fast simulation. On a PC with an Nvidia 2080Ti, we can achieve up to 1000\,fps when rendering single tactile images at 128$\times$128-resolution. Multiple PyBullet physics engines can also run in parallel, so during training we used 10 vectorised environments to increase throughput. 
    
\textbf{Flexible:} Whilst this study is predominantly based on the TacTip sensor, we do not attempt to simulate images accurate to any specific tactile sensor. Instead, we simulate only useful tactile features, relying on a later image-translation stage to map from real to simulated images. We expect similar results are possible with a broad range of other high-resolution optical tactile sensors, including sensors of the Gelsight type \cite{Johnson2009RetrographicShape,padmanabha2020omnitact,lambeta2020digit}, the Soft-Bubble \cite{kuppuswamy2020soft} and optical shear-based sensors \cite{Sferrazza2019DesignSensor}. 
    
\textbf{Realistic:} Instead of aiming to realistically simulate the physical properties of any specific sensor, which is both difficult and computationally expensive, we simulate only the desired properties of an idealised tactile sensor, currently focussing only on contact geometry. Our data-driven approach helps bridge the sim-to-real gap, so there is no need to generate synthetic images that match images from a real sensor to high precision. 
    
\textbf{Ease of use:} The simulation suite has been open-sourced. The simulation only requires the commonly-used PyBullet, so our approach should have a low barrier of entry. Our approach should also extend readily to different tasks and other tactile sensors.

\vspace{-.5em}
\section{Reinforcement Learning Environments} \label{sec:rl_envs}

Each task is provided with a set of observation spaces to allow for verification of the environments, comparisons between tasks, and an examination of multi-modal visuotactile control. Four standard types of observation space are considered: 

\textbf{Env State:} Comprises state information from the simulator, which is difficult information to collect in the real world. We use this to give baseline performance for a task that is expected to act as an upper limit. The information in this state varies between environments but commonly includes tool center point (TCP) pose, TCP velocity, goal locations and the current state of the environment.

\begin{figure*}[t]
      \centering
      \scriptsize
      \begin{overpic}[width=1.0\linewidth]{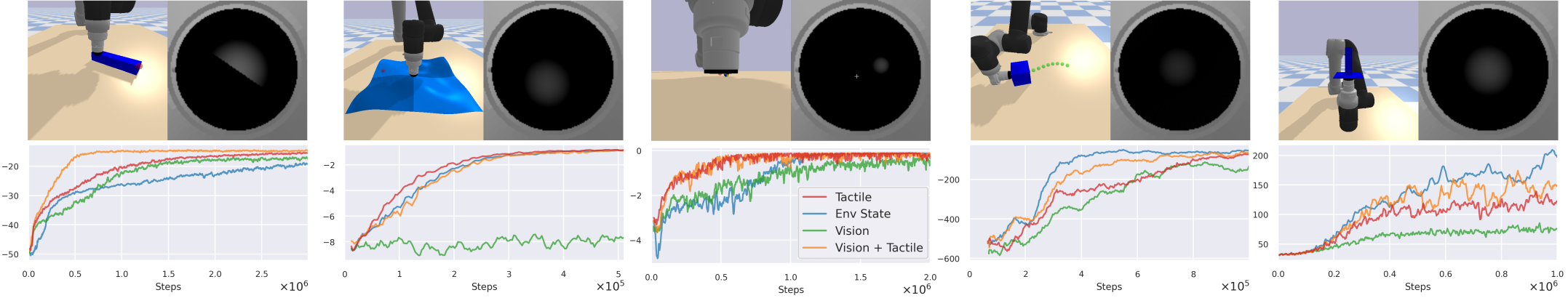}
      \put(-1, 19.5){\parbox{35mm}{\centering (a) Edge Following}}
      \put(19, 19.5){\parbox{35mm}{\centering (b) Surface Following}}
      \put(38, 19.5){\parbox{35mm}{\centering (c) Object Rolling}}
      \put(58, 19.5){\parbox{35mm}{\centering (d) Object Pushing}}
      \put(78,  19.5){\parbox{35mm}{\centering (e) Object Balancing}}
      \put(-1.25, 3){\rotatebox{90}{\tiny{Avg Rew}}}
      \end{overpic}
      \vspace{-1.5em}
      \caption{The DRL method PPO learns successful policies for the five considered RL environments. (a)~Traversing a randomly oriented edge while maintaining a set pose. (b)~Traversing a randomly-generated 3D surface while maintaining a set penetration and normal orientation. (c)~Manipulating a ball from a random initial position to a goal location. (d)~Manipulating a cube along a randomly-generated trajectory. (e)~Stabilising an object on the sensor tip. Panels show mean reward during training, smoothed with a window size of 50 and averaged over 3 seeds. Task reward commonly consists of negative distances between target and current poses, further details in Appendix \ref{apdx:RL_env_details}.}
      \vspace{0.5em}
      \label{fig:rl_envs_and_results}
    \end{figure*}
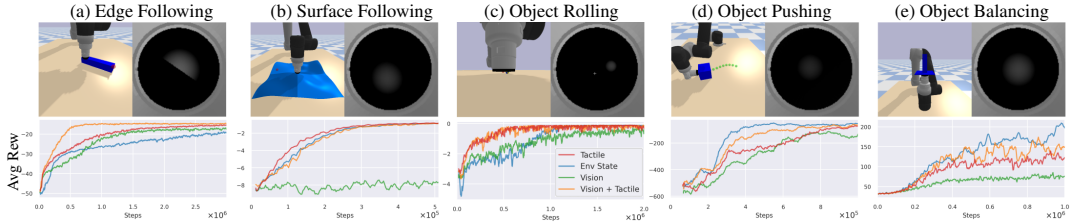

\textbf{Tactile:} Comprises images ($128\times128$) retrieved from the simulated optical tactile sensor attached to the end effector of the robot arm (Figure \ref{fig:rl_envs_and_results} right). Where tactile information alone is not sufficient to solve a task, this observation can be extended with state information retrieved from the simulator. This can only include information that could be easily and accurately captured in the real world, such as the TCP pose that is available on industrial robotic arms and the goal pose.
    
\textbf{Visual:} Comprises RGB images ($128\times128$) retrieved from a static, simulated camera viewing the environment (Figure \ref{fig:rl_envs_and_results} left). Only a single camera was used, although this could be extended to multiple cameras. As the simulated environment differs visually from the real-world environment, sim-to-real using RGB observations is challenging, requiring an approach like that of \cite{james2019sim, rao2020rl}.
    
\textbf{Visual + Tactile:} Combines the RGB visual and tactile image observations to into a 4-channel RGBT image. This case demonstrates a simple method of multi-modal sensing.  

\subsection{Tactile Exploration Environments} \label{subsec:sim_explore_envs}

In this set of tasks, the robot interacts with a physical environment that stays static, with the objective to learn policies to safely explore physical areas. These policies can be used to report information about novel objects, simplify human-control methods and improve operational safety of the robots. Two exploration environments have been created, for edge and surface following. More complete details on these environments are given in Tables \ref{tab:edge_follow_params} and \ref{tab:surface_follow_params} of Appendix \ref{apdx:RL_env_details}.

\textbf{Edge Following:} The edge-following environment is used to train an agent that maintains its pose relative to an edge whilst traversing the edge towards a goal location. Previous work has demonstrated that robust 2D contour following can be achieved with supervised learning techniques \cite{lepora2019pixels}. Here we demonstrate this task can also be completed via sim-to-real reinforcement learning. 

\textbf{Surface Following:} The surface-following environment is used to train an agent that maintains a set contact penetration of the sensor whilst orientating the TCP normal to an undulating 3D surface generated using OpenSimplex noise \cite{OpenSimplex}. For ease of use, we automatically direct the sensor in the direction of the goal, analogous to the pose-based tactile servo control methods introduced in \cite{lepora2020pose}. 

\subsection{Non-prehensile Manipulation Environments} \label{subsec:sim_manip_envs}
\vspace{-0.25em}

In this set of tasks, the objective is to learn policies that manipulate external objects in a desired manner. As our focus is on single sensors in this work, we consider non-prehensile manipulation. However, the simulation does support rendering of multiple tactile images, which could in principle be used simulate tactile grippers, manipulators or multiple robot arms in future work. Three manipulation tasks are proposed: object rolling, object pushing and object balancing. More details on the non-prehensile manipulation environments are given in Tables \ref{tab:object_rolling_params}, \ref{tab:object_pushing_params} and \ref{tab:object_balancing_params} of Appendix \ref{apdx:RL_env_details}.

\textbf{Object Rolling:} The object-rolling environment requires the manipulation of small spherical objects into a goal position within the TCP coordinate frame. The agent must learn a policy to roll the object from a random starting position to a random goal position. In this environment, a flat tactile sensor tip is used to simplify the motion needed to maintain contact with a spherical object. 

\textbf{Object Pushing:} In this task, the objective is to push an object to a goal location using the tactile image, as considered in previous work using supervised learning techniques \cite{lloyd2020goal}. Here we consider a cube pushed along a randomly-generated trajectory. The initial pose of the cube is randomised within limits and the trajectory generated using OpenSimplex noise. 

\textbf{Object Balancing:} This task is analogous to the well-known 2D inverted pendulum problem \cite{sutton-cartpole}, where an unstable pole with flat base is balanced on the tip of a sensor that points upwards. A random force perturbation is then applied to the object to cause instability. The objective is to learn a policy that applies planar actions to counteract the rotation of the balanced pole. 

\vspace{-0.25em}
\subsection{Reinforcement Learning Results} \label{subsubsec:reinforcement_learning}
\vspace{-0.25em}

This work uses the Stable-Baselines-3 \cite{stable-baselines3} implementation of PPO to train all tasks (further details in Appendix \ref{apdx:RL_details}). For an initial comparison, we train all tasks using all available observation spaces, averaged over 3 seeds. Training results are given in Figure~\ref{fig:rl_envs_and_results}, deterministic evaluation results and more detailed plots are given in Appendix \ref{apdx:full_RL_results}.

Environment state observations are used to verify that the environment leads to desired policies. As no image rendering is performed, this runs faster than tactile or visual observations; hence, it was used to find stable hyperparameters. Although comprised of ideal state information, we find that tactile image observations can lead to more sample-efficient and stable training, which is most clearly visible in the Edge-Following and Object-Rolling environments (Figs \ref{fig:rl_envs_and_results}a, \ref{fig:rl_envs_and_results}c).

Policies trained exclusively using visual observations tended to perform worst, which we attribute to most tasks being targeted towards tactile data; for example, details of the contact can be obscured using a single external camera. The best visual agent performance was for the object pushing task, which is the coarsest manipulation challenge. Whilst the object rolling task appears to perform well with visual observations, there is a notable performance gap where visual polices do not accurately learn the desired behaviour. When combining visual and tactile observations, successful learning can take place. Despite the additional complexity in the observations, agents can learn well with notable improvements in sample efficiency for the Edge-Follow and Object-Push tasks (Figs \ref{fig:rl_envs_and_results}a, \ref{fig:rl_envs_and_results}d). \looseness=-1

\vspace{-0.25em}
\section{Real-to-Sim Image Translation} \label{sec:real2sim_image_trans}
\vspace{-0.25em}

An advantage of using tactile images as the main form of observation is that the image space is more simplistic (e.g. markers on a uniform background) compared with the variety of visual images from a scene. Vision-based images are affected by features such as changing lighting conditions, shadows and texture that usually constitute a superfluous level of detail when solving a task. Simulating these visual features can make the learning more difficult because of the increased complexity of the observations and the added computation to render these images; conversely, ignoring these visual features would make image translation more difficult because of the greater simulation and reality gap, as they are still prevalent in reality, necessitating techniques such as image randomisation or RL task-aware training \cite{james2019sim, rao2020rl}. These complexities are not present for the real and simulated tactile images considered here because the camera is confined within the enclosure of the tactile sensor.

Conversely, a disadvantage of using tactile images as the main observations is that the tasks require the robot to be in close proximity, or touching, the environment. Whilst this contact may be necessary for some tasks, such as manipulation, it does make exploration in reality a challenge because of the self-inflicted damage that can occur. The approach proposed here is to use a separate data-collection stage, where the tactile sensor explores a series of configurations in a safe and controlled manner. As the tactile image space is relatively confined, an efficient exploration of a representative sample of configurations for a specific RL task becomes possible. Whilst some sensor configurations are not possible to sample in reality, such as large penetrations of the sensor that would cause damage, we aim to train a model that generalises to those unreachable configurations. 

Specifically, in this work we treat the sim-to-real problem as a supervised image-to-image translation problem. The same data collection procedure is performed in both simulation and reality, producing a data set of real and simulated image pairs. Here we choose to transfer from real-to-sim images, because the real images are richer in information with details such as shear forces that we choose not to model in simulation. Generative Adversarial Networks (GANs) are the state of the art for realistic image generation. Here we use the pix2pix \cite{pix2pix2017} architecture for image-to-image translation, which uses the U-net \cite{ronneberger2015u} architecture for the image conditioned generator and a standard convolutional network for the discriminator (Appendix \ref{apdx:pix2pix_arch} shows this architecture applied to tactile images). 

\vspace{-0.25em}
\subsection{Data Collection} \label{subsec:data_collection}
\vspace{-0.25em}

Three data sets corresponding to distinct environments are collected for sim-to-real transfer, each with 5000 image pairs for training and 2000 pairs for validation. We approximate the task environment with a simplified static environment consisting of 3d-printed components and no specialist hardware. Per task, these datasets take $\sim$5 hours to collect on physical hardware and $\sim$100 seconds in simulation. Following a previous investigation of contact-induced shear with the same optical tactile sensor \cite{LeporaOptimalTouchRAM}, we deliberately induce shear perturbations by randomly sliding the sensor during data collection (details given in Appendix \ref{apdx:pix2pix}). Past work has shown this step is key to ensuring that the trained neural network outputs are insensitive to unavoidable motion-dependent shear during task performance (for more details, see \cite{lepora2020pose}). We do not model any sensor shear in simulation to ensure that the real-to-sim image generation is also insensitive to shear on the real sensor.  

\vspace{-0.25em}
\subsection{Pix2Pix Training} \label{subsec:pix2pix}
\vspace{-0.25em}

The only change to the pix2pix architecture was to replace instance normalisation with spectral normalisation, with other default parameters sufficient for training on tactile images (full details in Appendix \ref{apdx:pix2pix_arch}). This change reduced droplet artefacts that were otherwise prevalent in our generated images. In addition, excluding the border from the simulated tactile images improved training, as otherwise the GAN focused on generating a realistic border instead of the tactile imprint. This is important because the imprint is the useful component when learning control policies. Instead, the border is re-added from a saved reference image after generation occurs.

Accurate tactile image generation is achieved with minimal difference between generated and target images (example images for each data set shown in Appendix \ref{apdx:pix2pix}, Figure \ref{fig:gan_image_comparison}). For the edge, surface and probe datasets, mean SSIM scores across the full validation set are $0.9955$, $0.9924$ and $0.9942$ respectively. The source of the small errors are highlighted in the shown SSIM image differences (Figure \ref{fig:gan_image_comparison}, right). The imprint borders in the generated images lack some sharp details generated in simulation, which will likely be due to a slight elastic stretching of the skin of the real tip that is not modelled in simulation. An approach such as Gaussian smoothing of the simulated images, as in \cite{gomes2021generation}, could reduce this effect, although we did not find it necessary to explore in the present work.

Crucially, the generator can interpret and generate image features useful to training RL policies. For example, edge orientation and position are accurately captured in the GAN. Also, the generator can generalise to images unseen during training. For example, all training data only had imprints from one source in each training image; however, during inference, multiple sources can be applied and the generator still produces realistic outputs (video demonstrations available \href{https://sites.google.com/view/tactile-sim-to-real/home}{\emph{here}}). Similarly, the generator appears to generalise outside the training data for penetration depth, which is important for helping to ensure that the sensor can avoid damage by not pressing too far into a surface.

\vspace{-0.25em}
\subsection{Supervised Learning Comparison}
\vspace{-0.25em}

For comparison to the only existing sim-to-real approach for the TacTip~\cite{Ding2020Sim-to-RealSensing}, we implement a supervised learning task to predict the radial displacement and polar angle of an edge pressed into the sensor (results in Appendix \ref{apdx:supervised_comparison}). Domain adaptation with our proposed method of tactile simulation out-performs the previous approach using elastic deformation simulation and domain randomisation, with approximately a 3-fold improvement in Mean Absolute Error (MAE) of the predictions.

\begin{figure}
    \centering
    \begin{minipage}{.48\textwidth}
      \centering
      \scriptsize
      \begin{overpic}[width=1.0\linewidth]{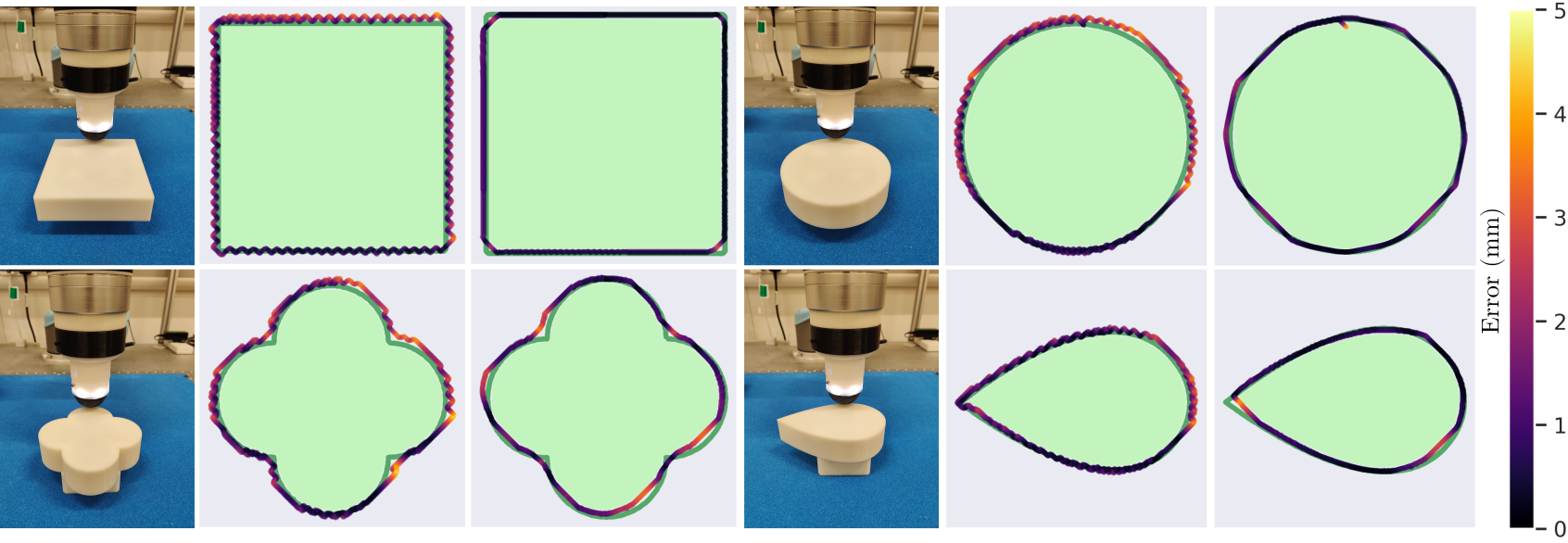}
        \put(6,   35.5){\parbox{20mm}{\centering Real}}
        \put(23,  35.5){\parbox{20mm}{\centering Sim}}        
        \put(54,  35.5){\parbox{20mm}{\centering Real}}
        \put(70,  35.5){\parbox{20mm}{\centering Sim}}
      \end{overpic}
      \captionof{figure}{Comparison between sim and real performance for the Edge-Following RL environment. Using several flat shapes (circle, square, clover, foil) to evaluate the policy generalisation performance.}
      \label{fig:real_edge_results}
    \end{minipage}
    \hfill
    \begin{minipage}{.48\textwidth}
      \centering
      \scriptsize
      \begin{overpic}[width=1.0\linewidth]{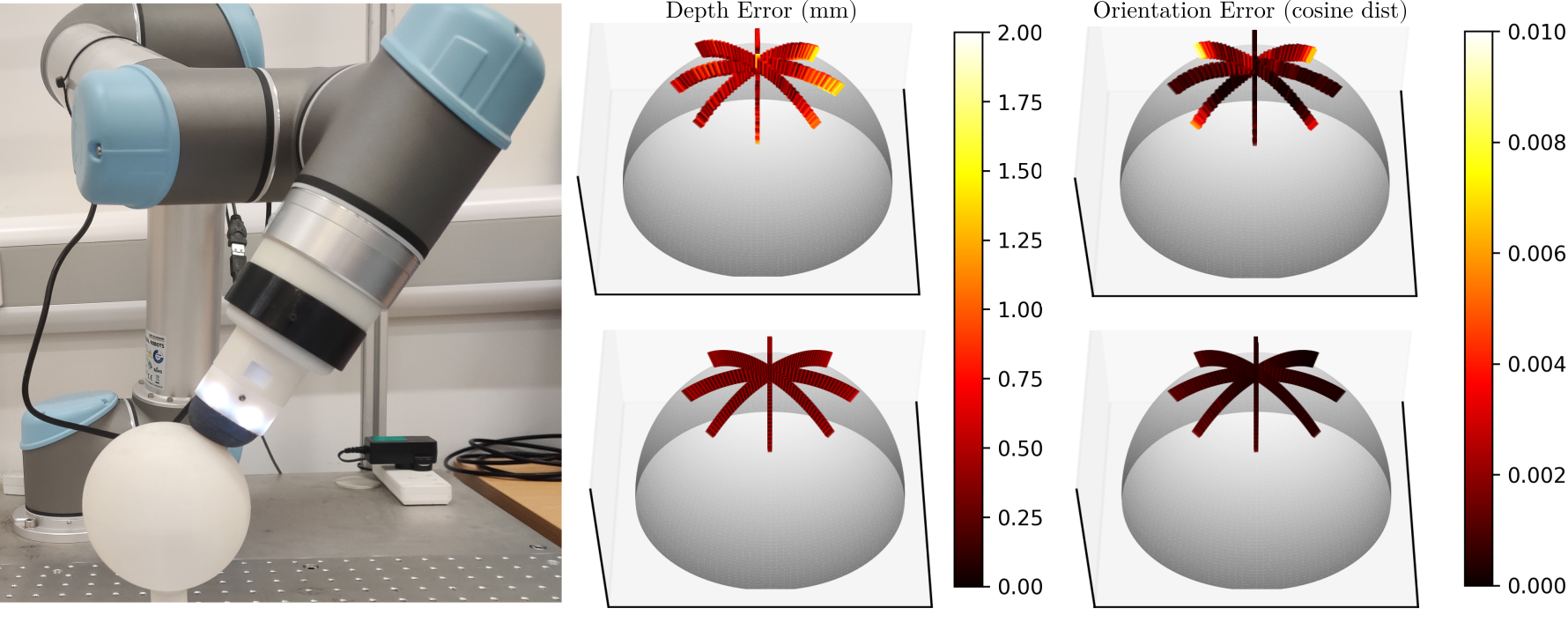}
          \put(27,   22){\parbox{20mm}{\centering Real}}
          \put(27,   2){\parbox{20mm}{\centering Sim}}
          \put(58,   22){\parbox{20mm}{\centering Real}}
          \put(58,   2){\parbox{20mm}{\centering Sim}}
      \end{overpic}
      \captionof{figure}{Comparison between sim and real performance for the Surface-Following RL environment. Positional error (left), Orientation error (right), Real data (top), simulated data (bottom).}
      \label{fig:real_surface_results}
    \end{minipage}
\end{figure}

\section{Results: Sim-to-Real Policy Transfer} \label{sec:sim2real_policy_trans}

Successful sim-to-real policy transfer was achieved on 4 of the 5 proposed tasks. Qualitative demonstration videos are available at \href{https://sites.google.com/my.bristol.ac.uk/tactile-gym-sim2real/home}{\emph{https://sites.google.com/my.bristol.ac.uk/tactile-gym-sim2real/home}}.

\begin{wraptable}{r}{0.4\linewidth}
\footnotesize
\caption{\label{tab:edge_quant_res}Quantitative results for edge following task. Mean distance from ground truth shape taken over 1000 evaluation steps.}
\vspace{-1.4em}
\begin{center}
\begin{tabular}{ c c c }
                 & \textbf{Real} & \textbf{Sim} \\ 
 \textbf{Square} & 1.47\si{\mm}  & 0.63\si{\mm} \\ 
 \textbf{Circle} & 1.50\si{\mm}  & 0.80\si{\mm} \\  
 \textbf{Clover} & 1.58\si{\mm}  & 1.38\si{\mm} \\  
 \textbf{Foil}   & 1.09\si{\mm}  & 0.47\si{\mm} 
\end{tabular}
\end{center}
\vspace{-0.5cm}
\end{wraptable}

\textbf{Edge Following:} The learned edge-following behaviour is evaluated over several novel flat shapes in the environment (Figure \ref{fig:real_edge_results}). The learned policy successfully traverses all objects, generalising to novel features such as curved edges and sharp internal or external corners not present during training. As these shapes are 3D printed, we have access to the ground truth in the CAD files (accurate to the precision of the 3D printer). We compare the trajectories taken in simulation to those in reality, reporting the distance from each point of the trajectory to the nearest ground truth node (Table \ref{tab:edge_quant_res}). The results show that the sensor maintains close proximity to the target edge.

The drop in performance on the physical robot seems related to an oscillating behaviour also present in simulation. The movements are exaggerated in reality due to increased latency between capturing an observation and predicting an action. A likely cause is the action-range clipping used in the PPO algorithm, which results in the action extremes being more prevalent. Approaches such as squashing functions \cite{haarnoja2018soft}, beta distributions \cite{chou2017beta} or constraint-based RL \cite{bohez2019value} could mitigate this artefact.

\begin{wraptable}{r}{0.4\linewidth}
\footnotesize
\vspace{0.5em}
\caption{Quantitative results for surface following task. Mean distance from ground truth evaluated over 1000 steps.}
\vspace{-1.5em}
\begin{center}
\begin{tabular}{ c c c }
                      & \textbf{Real} & \textbf{Sim} \\ 
 \textbf{Depth Error} & 0.57\si{\mm}  & 0.30\si{\mm} \\ 
 \textbf{Cosine Error} & 0.00118       & 0.00054
\end{tabular}
\end{center}
\vspace{-0.6cm}
\end{wraptable}

\textbf{Surface Following:} To evaluate the surface-following behaviour, a spherical object is traversed from its centre outwards in a set direction ranging over $360\si{\degree}$ in $45\si{\degree}$ intervals (Figure \ref{fig:real_surface_results}). As with the edge following task, we use a 3D-printed shape whose ground truth is known from the CAD model. The sensor successfully and accurately traverses the object on the physical robot despite texture and frictional forces not modelled in simulation. Like the edge following task, the accuracy drops between sim and real, although the evaluated policies still exhibit the desired behaviour. A more extensive qualitative test is given in the supplementary video for the undulating surface shown in Fig.\ref{fig:overview_full}.

\begin{figure}
    \centering
    \begin{minipage}{.48\textwidth}
        \centering
        \begin{overpic}[width=1.0\linewidth]{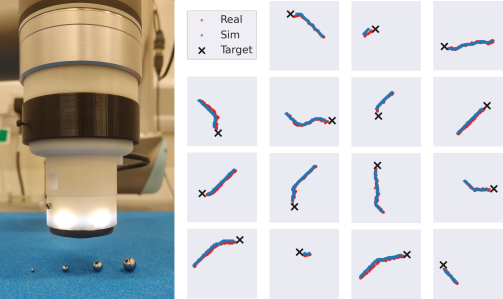}
        \end{overpic}
        \captionof{figure}{Real and simulated trajectories for policy evaluation of the object rolling task. A ball bearing is rolled an initial position to a goal location. Sizes range from 2-8\si{\mm} diameter (columns: left to right).}
        \label{fig:real_object_roll_results}
    \end{minipage}
    \hfill
    \begin{minipage}{.48\textwidth}
        \centering
        \scriptsize
        \begin{overpic}[width=1.0\linewidth]{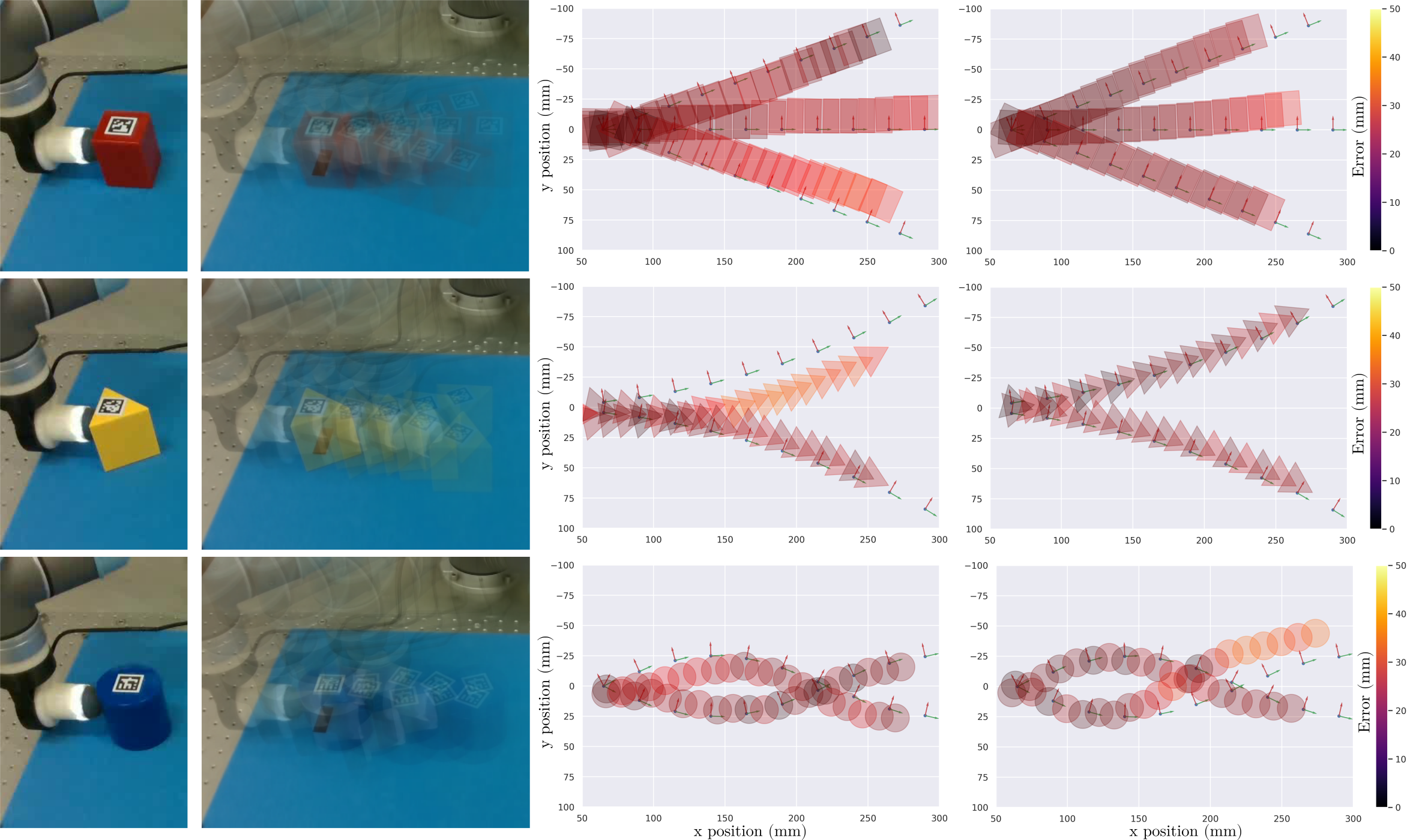}
          \put(38,   59.5){\parbox{20mm}{\centering Real}}
          \put(69,   59.5){\parbox{20mm}{\centering Sim}}
        \end{overpic}
  \captionof{figure}{Evaluation of the pushing task. Trajectories show a cube (top), triangular prism (middle) and cylinder (bottom) pushed along straight, curved and sinusoidal trajectories respectively.}
  \label{fig:real_object_push_results}
    \end{minipage}
    \vspace{.5em}
\end{figure}

\textbf{Object Rolling:} The object-rolling behaviour is tested by manipulating ball bearings (on an extended dimension range of $2$-$8\si{\mm}$ diameter) from random initial positions to random goal locations. During the evaluation, the position of the ball bearing relative to the sensor is obscured, so the imprint of the object is tracked on the tactile image using basic computer vision (blob detection). The target is reached when the pixel distance is less than a 5-pixel threshold ($\sim$2\si{\mm}). On the physical robot, this is performed for 25 trajectories with each bearing, resulting in 100 consecutive successful trials (subset shown in Figure~\ref{fig:real_object_roll_results}). To compare with simulated results, we apply the same initial and target positions then perform the same pixel tracking.
\begin{wraptable}{r}{0.5\linewidth}
\footnotesize
\vspace{0.75em}
\caption{Quantitative results for object pushing task. Mean Euclidean distance from trajectory.}
\vspace{-1.5em}
\begin{center}
\begin{tabular}{ c c c c }
                    & \textbf{Straight} & \textbf{Curve} & \textbf{Sine} \\ \hline
\textbf{Real Cube}  & 11.5\si{\mm}  & 11.7\si{\mm} & 13.6\si{\mm}  \\ 
\textbf{Sim Cube}   & 11.1\si{\mm}  & 10.1\si{\mm} & 12.7\si{\mm}  \\ \hline
\textbf{Real Tri}   & 11.0\si{\mm}  & 14.0\si{\mm} & 12.6\si{\mm}  \\ 
\textbf{Sim Tri}    & 21.8\si{\mm}  & 10.2\si{\mm} & 11.0\si{\mm}  \\ \hline
\textbf{Real Cyl}   & 13.3\si{\mm}  & 16.7\si{\mm} & 9.9\si{\mm}  \\ 
\textbf{Sim Cyl}    & 24.1\si{\mm}  & 13.9\si{\mm} & 12.7\si{\mm}  
\end{tabular}
\end{center}
\vspace{-2.5em}
\end{wraptable}
The sim and real results are similar, albeit with greater noise in the real trajectories.

\textbf{Object Pushing:} To evaluate this task we tracked the objects following the method given in \cite{lloyd2020goal}. Importantly Aruco markers were only used for obtaining quantitative results and not as part of the observation. This task was the most challenging to achieve accurate performance, likely due to the approximations of the simulated sensor tip, object being manipulated, and frictional interactions. Despite this, when evaluated in reality, task success was achieved across several trajectories with objects of different shape to that seen during training (examples in Figure \ref{fig:real_object_push_results}). Performance both in simulation and reality was sensitive to initial conditions and perturbations, with compounding errors causing the object to veer off the trajectory. Future work could improve robustness by learning under additional dynamics randomisation, introduction of random force perturbations or by better matching the physical properties of the real objects.

\textbf{Object Balancing:} The physical task remains an outstanding challenge, because the successful training in simulation required physics parameters outside the capabilities of our UR5 robot. 

\vspace{-0.25em}
\section{Discussion and Future Work}\label{sec:conclusion}
\vspace{-0.25em}
In this work, we demonstrated that zero-shot sim-to-real policy transfer is a viable approach for tactile-based RL agents. To learn the policy, we created a fast method for simulating tactile images based on contact geometry. From these images, distinct policies were learned for several physically-interactive tasks requiring a sense of touch. We demonstrated that a data-driven model for real-to-sim image translation can be embedded into the control loop for successful sim-to-real policy transfer. There are several future directions for extending and improving this approach.

The tactile simulation of both the contact dynamics and captured information in the model could be improved. Our contact dynamics model used used rigid bodies with soft contacts as an approximation to the deformation of real tactile sensors. Extending this work with soft-body deformation should enable a more realistic simulation, which was not pursued in this initial study because of increased computational costs and instability issues with the simulators. The captured information in the present study focused exclusively on contact geometry, which in principle could be extended to include global shear information through contact data available in most physics simulators. Texture could be also included using rendering techniques common in photo-realistic image generation. Local shear of the sensor would be challenging to simulate, for example during incipient slip, due to the contact reduction commonly used in physics engines for computational efficiency; however, this capability may be possible in soft-body simulations where these local forces are required.

For translating real to simulated images, improvement could be made to both the GAN and training data. We used a conventional pix2pix \cite{pix2pix2017} approach, which can be improved with extended methods \cite{wang2018pix2pixHD}. In addition, a distinct data set was collected for training the GAN in each learning environment. Because of the constrained nature of the tactile image space, a general tactile data set could potentially be used train a single image-translation model for policy transfer over multiple tasks. 

The environments considered here focused on skills achievable with only a single sensor and where the optimal behaviour is obvious for a human to interpret, as needed to verify that RL with a sim-to-real approach is a viable method. Future work could focus on tasks where ideal control policies are more complicated or unknown, so the RL framework can be fully exploited. An interesting topic would be to extend these methods to prehensile tasks such as grasping and dexterous manipulation, where tactile information will be valuable for learning desirable and robust policies. Although only one sensor was used in this current work, our tactile simulation does support multiple sensors, offering the opportunity to extend to more complex tasks involving hands with multiple tactile fingertips.

\clearpage

\section*{Acknowledgements}
Alex Church was supported by an EPSRC CASE award sponsored by Google DeepMind. John Lloyd and Nathan Lepora were supported by a Research Leadership Award from the Leverhulme Trust on ‘A biomimetic forebrain for robot touch’ (RL-2016-39). 

\setlength{\bibsep}{2.5pt}
\bibliography{references}  

\begin{thebibliography}{45}
\providecommand{\natexlab}[1]{#1}
\providecommand{\url}[1]{\texttt{#1}}
\expandafter\ifx\csname urlstyle\endcsname\relax
  \providecommand{\doi}[1]{doi: #1}\else
  \providecommand{\doi}{doi: \begingroup \urlstyle{rm}\Url}\fi

\bibitem[Akkaya et~al.(2019)Akkaya, Andrychowicz, Chociej, Litwin, McGrew,
  Petron, Paino, Plappert, Powell, Ribas, et~al.]{akkaya2019solving}
I.~Akkaya, M.~Andrychowicz, M.~Chociej, M.~Litwin, B.~McGrew, A.~Petron,
  A.~Paino, M.~Plappert, G.~Powell, R.~Ribas, et~al.
\newblock Solving rubik's cube with a robot hand.
\newblock \emph{arXiv:1910.07113}, 2019.

\bibitem[Lee et~al.(2020)Lee, Hwangbo, Wellhausen, Koltun, and
  Hutter]{Leeeabc5986}
J.~Lee, J.~Hwangbo, L.~Wellhausen, V.~Koltun, and M.~Hutter.
\newblock Learning quadrupedal locomotion over challenging terrain.
\newblock \emph{Science Robotics}, 5\penalty0 (47), 2020.

\bibitem[Ibarz et~al.(2021)Ibarz, Tan, Finn, Kalakrishnan, Pastor, and
  Levine]{IbarzRobotTrain}
J.~Ibarz, J.~Tan, C.~Finn, M.~Kalakrishnan, P.~Pastor, and S.~Levine.
\newblock How to train your robot with deep reinforcement learning: lessons we
  have learned.
\newblock \emph{IJRR}, 2021.

\bibitem[Schulman et~al.(2017)Schulman, Wolski, Dhariwal, Radford, and
  Klimov]{Schulman2017ProximalAlgorithms}
J.~Schulman, F.~Wolski, P.~Dhariwal, A.~Radford, and O.~Klimov.
\newblock Proximal policy optimization algorithms.
\newblock \emph{arXiv:1707.06347}, 2017.

\bibitem[James et~al.(2019)James, Wohlhart, Kalakrishnan, Kalashnikov, Irpan,
  Ibarz, Levine, Hadsell, and Bousmalis]{james2019sim}
S.~James, P.~Wohlhart, M.~Kalakrishnan, D.~Kalashnikov, A.~Irpan, J.~Ibarz,
  S.~Levine, R.~Hadsell, and K.~Bousmalis.
\newblock Sim-to-real via sim-to-sim: Data-efficient robotic grasping via
  randomized-to-canonical adaptation networks.
\newblock In \emph{CVPR}, pages 12627--12637, 2019.

\bibitem[Narang et~al.(2021{\natexlab{a}})Narang, Sundaralingam, Macklin,
  Mousavian, and Fox]{narang2021sim}
Y.~Narang, B.~Sundaralingam, M.~Macklin, A.~Mousavian, and D.~Fox.
\newblock Sim-to-real for robotic tactile sensing via physics-based simulation
  and learned latent projections.
\newblock \emph{arXiv preprint arXiv:2103.16747}, 2021{\natexlab{a}}.

\bibitem[Narang et~al.(2021{\natexlab{b}})Narang, Sundaralingam, Van~Wyk,
  Mousavian, and Fox]{narang2021interpreting}
Y.~S. Narang, B.~Sundaralingam, K.~Van~Wyk, A.~Mousavian, and D.~Fox.
\newblock Interpreting and predicting tactile signals for the syntouch biotac.
\newblock \emph{arXiv preprint arXiv:2101.05452}, 2021{\natexlab{b}}.

\bibitem[Sferrazza et~al.(2020)Sferrazza, Bi, and
  D’Andrea]{sferrazza2020learning}
C.~Sferrazza, T.~Bi, and R.~D’Andrea.
\newblock Learning the sense of touch in simulation: a sim-to-real strategy for
  vision-based tactile sensing.
\newblock In \emph{2020 IEEE/RSJ International Conference on Intelligent Robots
  and Systems (IROS)}, pages 4389--4396. IEEE, 2020.

\bibitem[Sferrazza and D'Andrea(2020)]{sferrazza2020sim}
C.~Sferrazza and R.~D'Andrea.
\newblock Sim-to-real for high-resolution optical tactile sensing: From images
  to 3d contact force distributions.
\newblock \emph{arXiv preprint arXiv:2012.11295}, 2020.

\bibitem[Bi et~al.(2021)Bi, Sferrazza, and D'Andrea]{bi2021zero}
T.~Bi, C.~Sferrazza, and R.~D'Andrea.
\newblock Zero-shot sim-to-real transfer of tactile control policies for
  aggressive swing-up manipulation.
\newblock \emph{IEEE Robotics and Automation Letters}, 2021.

\bibitem[Ding et~al.(2020)Ding, Lepora, and Johns]{Ding2020Sim-to-RealSensing}
Z.~Ding, N.~F. Lepora, and E.~Johns.
\newblock {Sim-to-Real Transfer for Optical Tactile Sensing}.
\newblock \emph{ICRA}, pages 1639--1645, 2020.

\bibitem[Ward-Cherrier et~al.(2018)Ward-Cherrier, Pestell, Cramphorn, Winstone,
  Giannaccini, Rossiter, and Lepora]{Ward-Cherrier2018}
B.~Ward-Cherrier, N.~Pestell, L.~Cramphorn, B.~Winstone, M.~E. Giannaccini,
  J.~Rossiter, and N.~F. Lepora.
\newblock {The TacTip Family: Soft Optical Tactile Sensors with 3D-Printed
  Biomimetic Morphologies}.
\newblock \emph{Soft Robotics}, 5\penalty0 (2):\penalty0 216--227, 2018.
\newblock ISSN 2169-5172.

\bibitem[Lepora(2021)]{lepora2021}
N.~F. Lepora.
\newblock Soft biomimetic optical tactile sensing with the tactip: A review.
\newblock \emph{IEEE Sensors Journal}, 21\penalty0 (19):\penalty0 21131--21143,
  2021.

\bibitem[Gomes et~al.(2021)Gomes, Paoletti, and Luo]{gomes2021generation}
D.~F. Gomes, P.~Paoletti, and S.~Luo.
\newblock Generation of gelsight tactile images for sim2real learning.
\newblock \emph{IEEE Robotics and Automation Letters}, 6\penalty0 (2):\penalty0
  4177--4184, 2021.

\bibitem[Wang et~al.(2020)Wang, Lambeta, Chou, and Calandra]{Wang2020TACTO}
S.~Wang, M.~Lambeta, L.~Chou, and R.~Calandra.
\newblock Tacto: A fast, flexible and open-source simulator for high-resolution
  vision-based tactile sensors.
\newblock \emph{arxiv:2012.08456}, 2020.

\bibitem[Johnson and Adelson(2009)]{Johnson2009RetrographicShape}
M.~K. Johnson and E.~H. Adelson.
\newblock Retrographic sensing for the measurement of surface texture and
  shape.
\newblock In \emph{CVPR}, pages 1070--1077. IEEE, 2009.

\bibitem[Padmanabha et~al.(2020)Padmanabha, Ebert, Tian, Calandra, Finn, and
  Levine]{padmanabha2020omnitact}
A.~Padmanabha, F.~Ebert, S.~Tian, R.~Calandra, C.~Finn, and S.~Levine.
\newblock Omnitact: A multi-directional high-resolution touch sensor.
\newblock In \emph{ICRA}, pages 618--624. IEEE, 2020.

\bibitem[Lambeta et~al.(2020)Lambeta, Chou, Tian, Yang, Maloon, Most, Stroud,
  Santos, Byagowi, Kammerer, et~al.]{lambeta2020digit}
M.~Lambeta, P.-W. Chou, S.~Tian, B.~Yang, B.~Maloon, V.~R. Most, D.~Stroud,
  R.~Santos, A.~Byagowi, G.~Kammerer, et~al.
\newblock Digit: A novel design for a low-cost compact high-resolution tactile
  sensor with application to in-hand manipulation.
\newblock \emph{RAL}, 5\penalty0 (3):\penalty0 3838--3845, 2020.

\bibitem[Kuppuswamy et~al.(2020)Kuppuswamy, Alspach, Uttamchandani, Creasey,
  Ikeda, and Tedrake]{kuppuswamy2020soft}
N.~Kuppuswamy, A.~Alspach, A.~Uttamchandani, S.~Creasey, T.~Ikeda, and
  R.~Tedrake.
\newblock Soft-bubble grippers for robust and perceptive manipulation.
\newblock \emph{arXiv:2004.03691}, 2020.

\bibitem[Sferrazza and D’Andrea(2019)]{Sferrazza2019DesignSensor}
C.~Sferrazza and R.~D’Andrea.
\newblock {Design, Motivation and Evaluation of a Full-Resolution Optical
  Tactile Sensor}.
\newblock \emph{Sensors}, 19\penalty0 (4):\penalty0 928, 2019.
\newblock ISSN 1424-8220.

\bibitem[Rao et~al.(2020)Rao, Harris, Irpan, Levine, Ibarz, and
  Khansari]{rao2020rl}
K.~Rao, C.~Harris, A.~Irpan, S.~Levine, J.~Ibarz, and M.~Khansari.
\newblock Rl-cyclegan: Reinforcement learning aware simulation-to-real.
\newblock In \emph{CVPR}, pages 11154--11163. {IEEE}, 2020.

\bibitem[Lepora et~al.(2019)Lepora, Church, De~Kerckhove, Hadsell, and
  Lloyd]{lepora2019pixels}
N.~F. Lepora, A.~Church, C.~De~Kerckhove, R.~Hadsell, and J.~Lloyd.
\newblock From pixels to percepts: Highly robust edge perception and contour
  following using deep learning and an optical biomimetic tactile sensor.
\newblock \emph{IEEE Robotics and Automation Letters}, 4\penalty0 (2):\penalty0
  2101--2107, 2019.

\bibitem[Kurt and A(2021)]{OpenSimplex}
S.~Kurt and S.~A.
\newblock Opensimplex noise.
\newblock \url{https://github.com/lmas/opensimplex}, 2021.

\bibitem[Lepora and Lloyd(2021)]{lepora2020pose}
N.~F. Lepora and J.~Lloyd.
\newblock Pose-based tactile servoing: Controlled soft touch using deep
  learning.
\newblock \emph{IEEE Robotics Automation Magazine}, pages 2--15, 2021.

\bibitem[Lloyd and Lepora(2021)]{lloyd2020goal}
J.~Lloyd and N.~F. Lepora.
\newblock Goal-driven robotic pushing using tactile and proprioceptive
  feedback.
\newblock \emph{IEEE Transactions on Robotics}, pages 1--12, 2021.

\bibitem[{Barto} et~al.(1983){Barto}, {Sutton}, and
  {Anderson}]{sutton-cartpole}
A.~G. {Barto}, R.~S. {Sutton}, and C.~W. {Anderson}.
\newblock Neuronlike adaptive elements that can solve difficult learning
  control problems.
\newblock \emph{IEEE SMC}, 13\penalty0 (5):\penalty0 834--846, 1983.

\bibitem[Raffin et~al.(2019)Raffin, Hill, Ernestus, Gleave, Kanervisto, and
  Dormann]{stable-baselines3}
A.~Raffin, A.~Hill, M.~Ernestus, A.~Gleave, A.~Kanervisto, and N.~Dormann.
\newblock Stable baselines3.
\newblock Available: \url{https://github.com/DLR-RM/stable-baselines3}, 2019.

\bibitem[Isola et~al.(2017)Isola, Zhu, Zhou, and Efros]{pix2pix2017}
P.~Isola, J.-Y. Zhu, T.~Zhou, and A.~A. Efros.
\newblock Image-to-image translation with conditional adversarial networks.
\newblock \emph{CVPR}, 2017.

\bibitem[Ronneberger et~al.(2015)Ronneberger, Fischer, and
  Brox]{ronneberger2015u}
O.~Ronneberger, P.~Fischer, and T.~Brox.
\newblock U-net: Convolutional networks for biomedical image segmentation.
\newblock In \emph{MICCAI}, pages 234--241. Springer, 2015.

\bibitem[Lepora and Lloyd(2020)]{LeporaOptimalTouchRAM}
N.~F. Lepora and J.~Lloyd.
\newblock Optimal deep learning for robot touch: Training accurate pose models
  of 3d surfaces and edges.
\newblock \emph{IEEE Robotics Automation Magazine}, 27\penalty0 (2):\penalty0
  66--77, 2020.

\bibitem[Haarnoja et~al.(2018)Haarnoja, Zhou, Hartikainen, Tucker, Ha, Tan,
  Kumar, Zhu, Gupta, Abbeel, et~al.]{haarnoja2018soft}
T.~Haarnoja, A.~Zhou, K.~Hartikainen, G.~Tucker, S.~Ha, J.~Tan, V.~Kumar,
  H.~Zhu, A.~Gupta, P.~Abbeel, et~al.
\newblock Soft actor-critic algorithms and applications.
\newblock \emph{arXiv:1812.05905}, 2018.

\bibitem[Chou(2017)]{chou2017beta}
P.-W. Chou.
\newblock The beta policy for continuous control reinforcement learning.
\newblock Master's thesis, Pittsburgh: Carnegie Mellon University, 2017.

\bibitem[Bohez et~al.(2019)Bohez, Abdolmaleki, Neunert, Buchli, Heess, and
  Hadsell]{bohez2019value}
S.~Bohez, A.~Abdolmaleki, M.~Neunert, J.~Buchli, N.~Heess, and R.~Hadsell.
\newblock Value constrained model-free continuous control.
\newblock \emph{arXiv:1902.04623}, 2019.

\bibitem[Wang et~al.(2018)Wang, Liu, Zhu, Tao, Kautz, and
  Catanzaro]{wang2018pix2pixHD}
T.~Wang, M.~Liu, J.~Zhu, A.~Tao, J.~Kautz, and B.~Catanzaro.
\newblock High-resolution image synthesis and semantic manipulation with
  conditional gans.
\newblock In \emph{CVPR}, pages 8798--8807. IEEE, 2018.

\bibitem[Chen et~al.(2019)Chen, Jain, Schissler, Gari, Al-Halah, Ithapu,
  Robinson, and Grauman]{Chen2019SoundSpaces:Environments}
C.~Chen, U.~Jain, C.~Schissler, S.~V.~A. Gari, Z.~Al-Halah, V.~K. Ithapu,
  P.~Robinson, and K.~Grauman.
\newblock {SoundSpaces: Audio-Visual Navigation in 3D Environments}.
\newblock In \emph{ECCV}, volume 12351 LNCS, pages 17--36. Springer, 2019.

\bibitem[James et~al.(2019)James, Ma, Arrojo, and
  Davison]{James2019RLBench:Environment}
S.~James, Z.~Ma, D.~R. Arrojo, and A.~J. Davison.
\newblock {RLBench: The Robot Learning Benchmark {\&} Learning Environment}.
\newblock \emph{RAL}, 5\penalty0 (2):\penalty0 3019--3026, 2019.

\bibitem[Zhu et~al.(2020)Zhu, Wong, Mandlekar, and
  Mart{\'{i}}n-Mart{\'{i}}n]{Zhu2020Robosuite:Learning}
Y.~Zhu, J.~Wong, A.~Mandlekar, and R.~Mart{\'{i}}n-Mart{\'{i}}n.
\newblock {robosuite: A Modular Simulation Framework and Benchmark for Robot
  Learning}.
\newblock \emph{arXiv:2009.12293}, 2020.

\bibitem[Hu et~al.(2020)Hu, Anderson, Li, Sun, Carr, Ragan{-}Kelley, and
  Durand]{Hu2019DiffTaichi:Simulation}
Y.~Hu, L.~Anderson, T.~Li, Q.~Sun, N.~Carr, J.~Ragan{-}Kelley, and F.~Durand.
\newblock Difftaichi: Differentiable programming for physical simulation.
\newblock In \emph{Proc. of ICLR}, 2020.

\bibitem[Heiden et~al.(2020)Heiden, Millard, Coumans, Sheng, and
  Sukhatme]{Heiden2020NeuralSim:Networks}
E.~Heiden, D.~Millard, E.~Coumans, Y.~Sheng, and G.~S. Sukhatme.
\newblock {NeuralSim: Augmenting Differentiable Simulators with Neural
  Networks}.
\newblock \emph{arXiv:2011.04217}, 2020.

\bibitem[Coumans and Bai(2016--2019)]{coumans2019}
E.~Coumans and Y.~Bai.
\newblock Pybullet, a python module for physics simulation for games, robotics
  and machine learning.
\newblock Available: \url{http://pybullet.org}, 2016--2019.

\bibitem[Matas et~al.(2018)Matas, James, and
  Davison]{Matas2018Sim-to-RealManipulation}
J.~Matas, S.~James, and A.~J. Davison.
\newblock Sim-to-real reinforcement learning for deformable object
  manipulation.
\newblock In \emph{CoRL}, pages 734--743. PMLR, 2018.

\bibitem[Mnih et~al.(2015)Mnih, Kavukcuoglu, Silver, Rusu, Veness, Bellemare,
  Graves, Riedmiller, Fidjeland, Ostrovski, et~al.]{mnih2015human}
V.~Mnih, K.~Kavukcuoglu, D.~Silver, A.~A. Rusu, J.~Veness, M.~G. Bellemare,
  A.~Graves, M.~Riedmiller, A.~K. Fidjeland, G.~Ostrovski, et~al.
\newblock Human-level control through deep reinforcement learning.
\newblock \emph{Nature}, 518\penalty0 (7540):\penalty0 529--533, 2015.

\bibitem[Kostrikov et~al.(2020)Kostrikov, Yarats, and
  Fergus]{kostrikov2020image}
I.~Kostrikov, D.~Yarats, and R.~Fergus.
\newblock Image augmentation is all you need: Regularizing deep reinforcement
  learning from pixels.
\newblock \emph{arXiv:2004.13649}, 2020.

\bibitem[Laskin et~al.(2020)Laskin, Lee, Stooke, Pinto, Abbeel, and
  Srinivas]{laskin2020reinforcement}
M.~Laskin, K.~Lee, A.~Stooke, L.~Pinto, P.~Abbeel, and A.~Srinivas.
\newblock Reinforcement learning with augmented data.
\newblock In \emph{Advances in Neural Information Processing Systems},
  volume~33, pages 19884--19895. Curran Associates, Inc., 2020.

\bibitem[Clevert et~al.(2015)Clevert, Unterthiner, and
  Hochreiter]{clevert2015fast}
D.-A. Clevert, T.~Unterthiner, and S.~Hochreiter.
\newblock Fast and accurate deep network learning by exponential linear units
  (elus).
\newblock \emph{arXiv preprint arXiv:1511.07289}, 2015.

\end{thebibliography}

\newpage
\appendix

\section{Robotics Simulation} \label{appdx:robot_sim}

\subsection{Physics Engine}

Physics-based simulation has played a vital role in the field of robotics, enabling rapid prototyping, testing and experimentation. The accuracy and speed of simulation has scaled with available computation. Modern methods such as RL have leveraged this to simulate very large data sets that are otherwise impractical to collect from the real world, hence far more complex and general behaviours can be learnt. The majority of simulated robotics has focused on rigid-body dynamics and vision sensors. More recently, a range of specific environment suites have been introduced to bring simulation closer to reality or to facilitate research in an under-represented direction \cite{Chen2019SoundSpaces:Environments, James2019RLBench:Environment, Zhu2020Robosuite:Learning, Hu2019DiffTaichi:Simulation, Heiden2020NeuralSim:Networks}. Collectively, these works highlight the importance of increasing the breadth and capabilities of simulation software available to researchers.

There are many choices for physics engines when developing a new suite. Here we choose the PyBullet~\cite{coumans2019} for its fast GPU rendering, support for deformable objects \cite{Matas2018Sim-to-RealManipulation}, fast and reliable kinematics and dynamics solvers, and its demonstrated sim-to-real success in robotics \cite{james2019sim}. Importantly, PyBullet is open source and non-commercialised software which helps to improve accessibility and lower the barrier of entry for RL research. That said, the tools used here to build our tactile suite are available in other physics engines such as Mujoco, Gazebo, Nvidia-isaacgym and Unity ML agents.

\subsection{Control}

Throughout this work we use Cartesian velocity control where the control input is a desired velocity (twist) specified by a 6 DoF action constrained to the allowed modes of control of the task. A control rate specifies the frequency that new actions can be sent to the robot, with maximum velocity limits also imposed. For most of our experiments, we set a velocity control rate of 10\,Hz, a maximum linear velocity of $\num{10}\,\si{\mm\per\second}$ and a maximum angular velocity of $\num{5}\,\si{\degree\per\second}$. When undergoing a series of predefined and random actions we notice no significant difference in Tool Centre Point (TCP) pose between simulation and the real robot.

Work-space coordinate frames are set in both simulation and reality, specific to each environment, with each action sent to the robot consisting of a Cartesian move relative to the work frame. 
\begin{wraptable}{r}{0.4\linewidth}
\footnotesize
\vspace{1em}
\caption{\label{tab:hyperparams} RL and network hyperparameters.}
\vspace{-1em}
\begin{center}
	\begin{tabular}{|c|l|c|c|c|}
		\hline
		& \multicolumn{1}{l|}{ \textbf{Param} } &  \multicolumn{3}{c|}{ \textbf{Value} }   \\ 
		\hline 
		
		\multirow{9}{*}{\STAB{\rotatebox[origin=c]{90}{ \textbf{Feature Extractor}}}}
		& Input dim        &  \multicolumn{3}{c|}{ [128, 128, $C$] }    \\
		& Conv filters     &  \multicolumn{3}{c|}{ [32, 64, 64]  }    \\
		& Kernel widths    &  \multicolumn{3}{c|}{ [8, 4, 3]     }    \\
		& Strides          &  \multicolumn{3}{c|}{ [4, 2, 1]     }    \\
		& Pooling          &  \multicolumn{3}{c|}{ None          }    \\
		& Output dim       &  \multicolumn{3}{c|}{ 512           }    \\
		& State Encoder    &  \multicolumn{3}{c|}{ [64, 64]      }    \\
		& Activation       &  \multicolumn{3}{c|}{ \textit{ReLU} }    \\ 
		& Initialiser      &  \multicolumn{3}{c|}{ \textit{Orthoganal} }        \\ 
		\hline 
		
		\multirow{3}{*}{\STAB{\rotatebox[origin=c]{90}{ \textbf{RL Net}}}}
		& Policy        &  \multicolumn{3}{c|}{ [256, 256] }         \\
		& Value         &  \multicolumn{3}{c|}{ [256, 256] }         \\
		& Activation    &  \multicolumn{3}{c|}{ \textit{Tanh} }         \\
		\hline

		\multirow{12}{*}{\STAB{\rotatebox[origin=c]{90}{ \textbf{PPO parameters} }}}
		& Learning Rate        &  \multicolumn{3}{c|}{ $3\times10^{-4}$ }  \\
		& n/ Envs              &  \multicolumn{3}{c|}{ 10 }  \\
		& Epoch steps          &  \multicolumn{3}{c|}{ 2048 }              \\
		& Batch size           &  \multicolumn{3}{c|}{ $64$ }              \\
		& n/ Epochs            &  \multicolumn{3}{c|}{ 10   }     \\
		& Discount ($\gamma$)  &  \multicolumn{3}{c|}{ 0.95 }     \\
		& GAE lambda           &  \multicolumn{3}{c|}{ 0.9  }     \\
		& Clip range           &  \multicolumn{3}{c|}{ 0.2  }     \\
		& Entropy coeff        &  \multicolumn{3}{c|}{ 0.0  }     \\ 
		& VF coeff             &  \multicolumn{3}{c|}{ 0.5  }     \\ 
		& Max grad norm        &  \multicolumn{3}{c|}{ 0.5  }     \\
		& KL limit             &  \multicolumn{3}{c|}{ 0.1  }     \\
		& Optimiser            &  \multicolumn{3}{c|}{ \textit{Adam} }     \\
		\hline  

	\end{tabular}
\end{center}
\vspace{-1.5cm}
\end{wraptable}
Therefore, the learned policies can be transferred from sim-to-real without exactly replicating the simulated task; for example edges and surfaces can be placed in alternative locations provided the work frame is set correctly. As a consequence, the policy transfers even when there are notable differences in the simulation, such as mirrored arm configurations used in this work. Thus, in principle the policies could also be transferred to other robot arms, providing the same speed and frequency of control can be achieved.

\section{Reinforcement Learning Parameters} \label{apdx:RL_details}

Near-default hyper-parameters are used in all training (full list in Table \ref{tab:hyperparams}). Image-based observations use the Atari Nature \cite{mnih2015human} convolutional layers followed by two 256-node fully connected (FC) layers. State observations use only the FC layers. For tasks that require both image and state data, the state data is passed through two 64-node FC layers and the output concatenated with the flattened output of the convolutional layers, which is then passed through the final FC layers for action and value prediction. The convolutional weights are shared for all policy and value networks. Small random image translation augmentations help to improve performance and stabilise training, as proposed in \cite{kostrikov2020image, laskin2020reinforcement}.

\newpage

\section{Reinforcement Learning Environment Details} \label{apdx:RL_env_details}

    \begin{table}[!htbp]
    \scriptsize
    \setlength{\tabcolsep}{1pt}
    
    \begin{minipage}{.475\linewidth}
    \centering
        \caption{\label{tab:edge_follow_params} Edge Follow environment description.}
        \begin{tabularx}{1.0\linewidth}{|l|X|}
        
        \hline
        \textbf{Observation}   & \tabitem \textbf{Env State:} \newline \{TCP pos, TCP lin vel, Goal pos, Edge ang\} \\
                              & \tabitem \textbf{Tactile:} \newline \{Tactile Image\} \\  
                              & \tabitem \textbf{Vision:} \newline \{RGB Image\}\\ 
                              & \tabitem \textbf{Vision + Tactile:} \newline \{RGB Image, Tactile Image\}\\  
        \hline
        
        \textbf{Action Space}  & $\{x, y\}$  \\ \hline
        
        \textbf{Reward}        & - ( Euclidean distance from TCP to goal + \\
                              & perpendicular distance from TCP to edge ) \\ \hline
        
        \textbf{Termination}   &  \tabitem max episode length reached \\
                              &  \tabitem euclidean distance from TCP to goal \textless{} 1\si{\cm}  \\ \hline
        
        \textbf{History}       & 1 Frame. \\ \hline
        
        \textbf{Randomisation} & \tabitem Edge randomly orientated through 360\si{\degree}. \\
                              & \tabitem Distance tip is embedded onto edge is randomly selected between  1.5\si{\mm} and 3.5\si{\mm}. 
        \\ \hline
        \end{tabularx}
    
    \end{minipage}\hfill
    \begin{minipage}{.475\linewidth}
        \centering
        \caption{\label{tab:surface_follow_params} Surface Follow environment description.}
    
        \begin{tabularx}{1.0\linewidth}{|l|X|}
        \hline
        \textbf{Observation}   & \tabitem \textbf{Env State:} \newline \{TCP pos, TCP orn, TCP lin vel, TCP ang vel, Goal pos, Target surface height, Target surface normal\} \\
                              & \tabitem \textbf{Tactile:} \newline \{Tactile Image\} \\  
                              & \tabitem \textbf{Vision:} \newline \{RGB Image\}\\ 
                              & \tabitem \textbf{Vision + Tactile:} \newline \{RGB Image, Tactile Image\}\\  
        \hline
        
        \textbf{Action Space}  & $\{z, Rx, Ry\}$  \\ \hline
        
        \textbf{Reward}        & -($z$ difference between TCP and local surface index + \\
                              & cosine difference between TCP normal and local surface normal) \\ \hline
        
        \textbf{Termination}   &  \tabitem max episode length reached \\
                              &  \tabitem euclidean distance from TCP to goal \textless{} 1\si{\cm} \\
                              \hline 
                               
        \textbf{History}       & 1 Frame. \\ \hline
        
        \textbf{Randomisation} & \tabitem Surface randomly generated w/ OpenSimplex noise. \\
                              & \tabitem Direction of goal randomly selected from $[0\si{\degree},360\si{\degree}]$.
        \\ \hline
        \end{tabularx}
    \end{minipage}\hfill
    \begin{minipage}{.475\linewidth}
        \centering
        \caption{\label{tab:object_rolling_params} Object Roll environment description.}
    
        \begin{tabularx}{1.0\linewidth}{|l|X|}
        \hline
        \textbf{Observation}   & \tabitem \textbf{Env State:} \newline \{TCP pos, TCP orn, TCP lin vel, TCP ang vel, Obj pos, Obj orn, Obj lin vel, Obj ang vel, Goal pos, Obj radius\} \\
                              & \tabitem \textbf{Tactile:} \newline \{Tactile Image, Goal pos\} \\  
                              & \tabitem \textbf{Vision:} \newline \{RGB Image, Goal pos\}\\ 
                              & \tabitem \textbf{Vision + Tactile:} \newline \{RGB Image, Tactile Image, Goal pos\}\\
                              \hline
                               
        \textbf{Action Space}  & $\{x, y\}$  \\ \hline
        
        \textbf{Reward}        & -(euclidean distance from object to goal) \\ \hline
        
        \textbf{Termination}   &  \tabitem max episode length reached \\
                              &  \tabitem euclidean distance from object to goal \textless{} 1\si{\mm} \\
                              \hline 
        
        \textbf{History} & 1 Frame. \\ \hline
        
        \textbf{Randomisation} & \tabitem Random starting position of object in TCP frame.\\
                              & \tabitem Random marble size between 5\si{\mm} and 10\si{\mm} diameter. \\
                              & \tabitem Random distance embedded into the sensor.
        \\ \hline
        \end{tabularx}
    \end{minipage}\hfill
    \begin{minipage}{.475\linewidth}
        \centering
        \caption{\label{tab:object_pushing_params} Object Push environment description.}
    
        \begin{tabularx}{1.0\linewidth}{|l|X|}
        \hline
        \textbf{Observation}   & \tabitem \textbf{Env State:} \newline \{TCP pos, TCP orn, TCP lin vel, TCP ang vel, Obj pos, Obj orn, Obj lin vel, Obj ang vel, Goal pos, Goal orn\} \\
                              & \tabitem \textbf{Tactile:} \newline \{Tactile Image, TCP pos, TCP orn, Goal pos, Goal orn\} \\  
                              & \tabitem \textbf{Vision:} \newline \{RGB Image, TCP pos, TCP orn, Goal pos, Goal orn\}\\ 
                              & \tabitem \textbf{Vision + Tactile:} \newline \{RGB Image, Tactile Image, TCP pos, TCP orn, Goal pos, Goal orn\}\\
                              \hline
        
        \textbf{Action Space}  & $\{y, Rz\}$  \\ \hline
        
        \textbf{Reward}        & -(Euclidean distance from object to goal + \\
                              & cosine distance from object orn to goal orn + \\ 
                              & cosine distance from TCP normal to object normal) \\ \hline
        
        \textbf{Termination}   &  \tabitem max episode length reached \\
                              &  \tabitem Euclidean distance from object to final goal \textless{} 2.5\si{\cm} \\
                              \hline 
        
        \textbf{History} & 1 Frame. \\ \hline
            
        \textbf{Randomisation} & \tabitem Random trajectory of goals generated with OpenSimplex Noise. 
        \\ \hline
        \end{tabularx}
    \end{minipage}\hfill
    \centering
    \begin{minipage}{.475\linewidth}
        \centering
        \caption{\label{tab:object_balancing_params} Object Balance environment description.}
    
        \begin{tabularx}{1.0\linewidth}{|l|X|}
        \hline
        \textbf{Observation}   & \tabitem \textbf{Env State:} \newline \{TCP pos, TCP orn, TCP lin vel, TCP ang vel, Obj pos, Obj orn, Obj lin vel, Obj ang vel\} \\
                              & \tabitem \textbf{Tactile:} \newline \{Tactile Image\} \\  
                              & \tabitem \textbf{Vision:} \newline \{RGB Image\}\\ 
                              & \tabitem \textbf{Vision + Tactile:} \newline \{RGB Image, Tactile Image\}\\
                              \hline
        
        \textbf{Action Space}  & $\{x, y\}$  \\ \hline
        
        \textbf{Reward}        & +1 per step \\ \hline
        
        \textbf{Termination}   & \tabitem max episode length reached \\
                              & \tabitem object tilts passed set angle ($35\si{\degree}$)  \\ \hline
        
        \textbf{History} & 2 Frames. \\ \hline
        
        \textbf{Randomisation} & \tabitem Random external force perturbation applied at start of episode. 
        \\ \hline
        \end{tabularx}
    \end{minipage}\hfill

    \end{table}

\newpage

\section{Full Reinforcement Learning Results} \label{apdx:full_RL_results}

  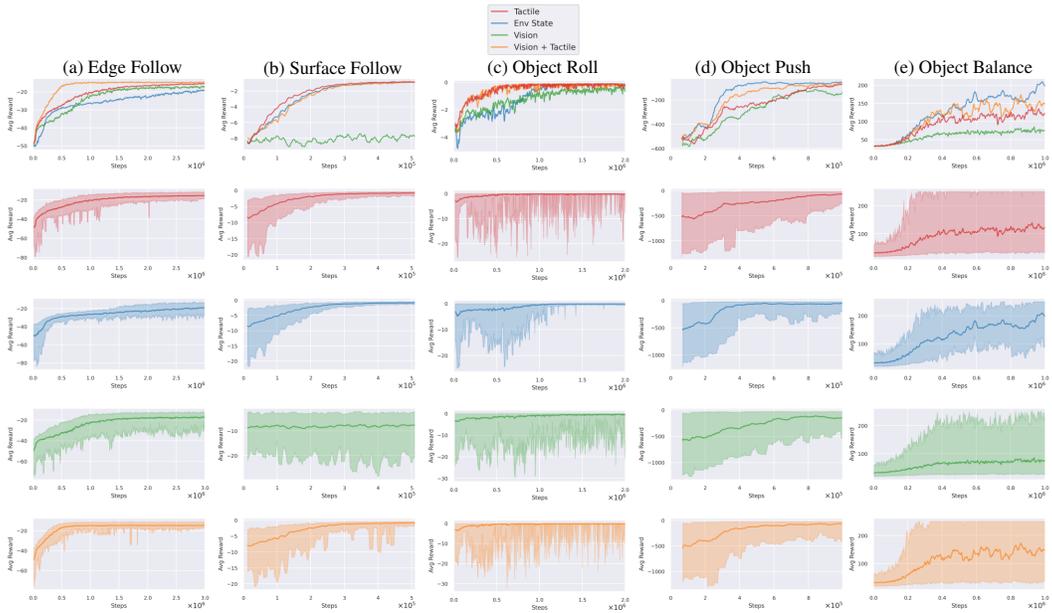
\begin{figure}[!htbp]
      \centering
      \scriptsize
      \begin{overpic}[width=1.0\linewidth]{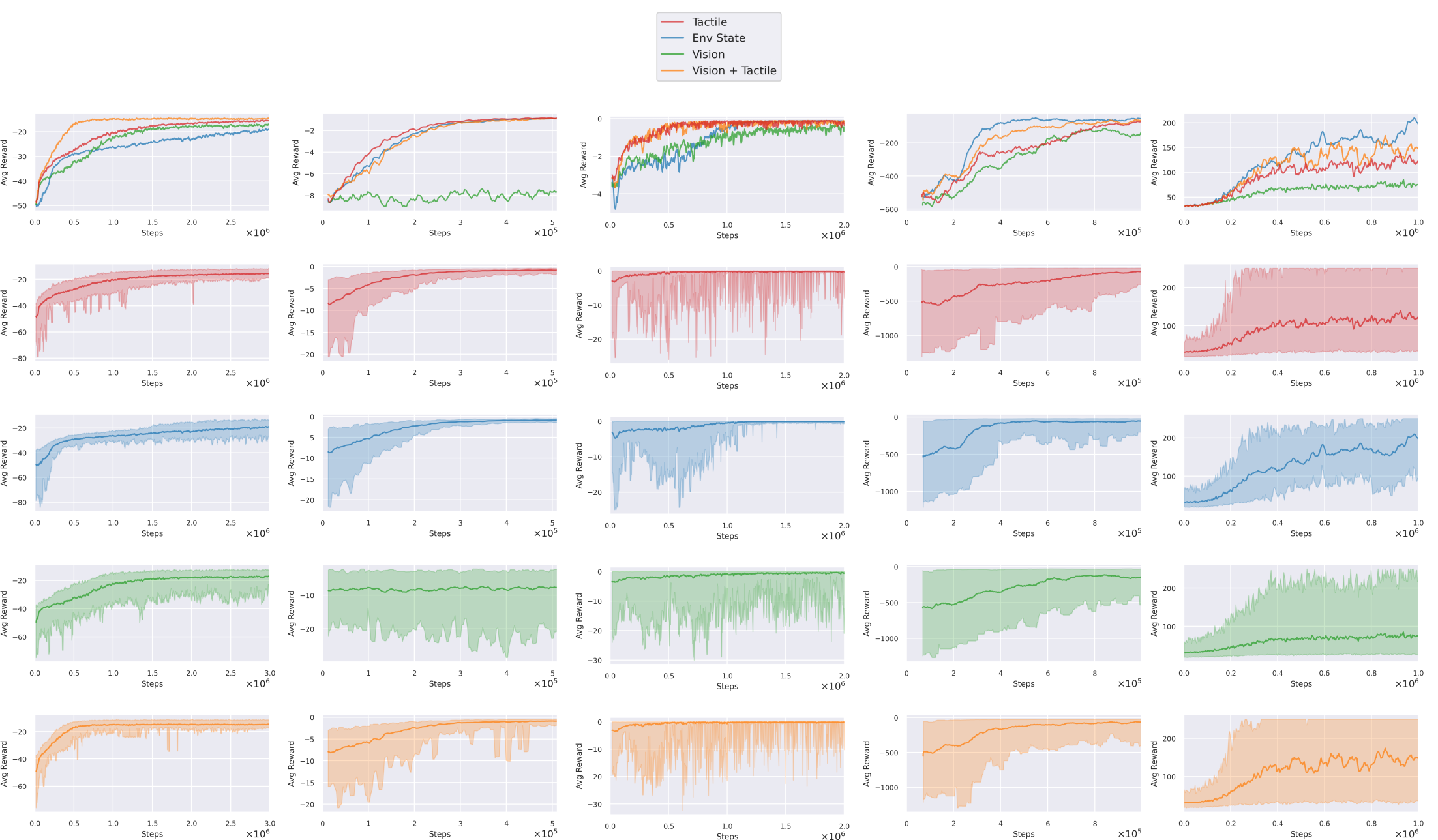}
          \put(2,  51){\parbox{25mm}{\centering (a) Edge Follow}}
          \put(22, 51){\parbox{25mm}{\centering (b) Surface Follow}}
          \put(42, 51){\parbox{25mm}{\centering (c) Object Roll}}
          \put(62, 51){\parbox{25mm}{\centering (d) Object Push}}
          \put(82, 51){\parbox{25mm}{\centering (e) Object Balance}}
      \end{overpic}
      \caption{Full training results of reinforcement learning agents. Results are smoothed with a window size of 50 followed by averaging over 3 seeds. Shaded regions indicate maximum and minimum reward achieved over the 3 seeds (after smoothing).}
      \label{fig:full_rl_train_results}
    \end{figure} 
    
    \vspace{2em}
    
    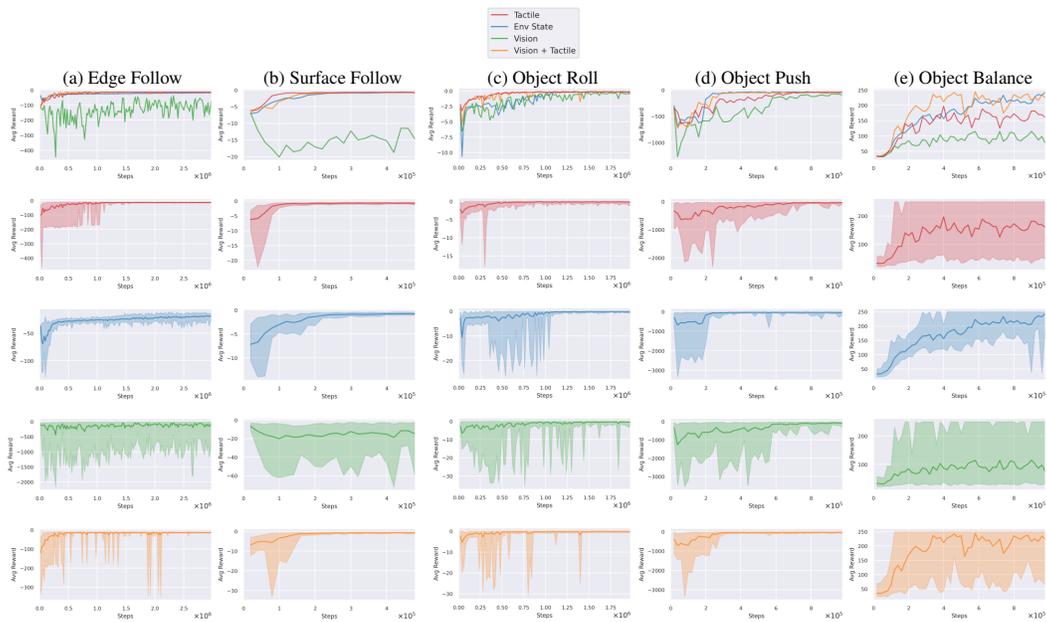
\begin{figure}[!htbp]
      \centering
      \scriptsize
      \begin{overpic}[width=1.0\linewidth]{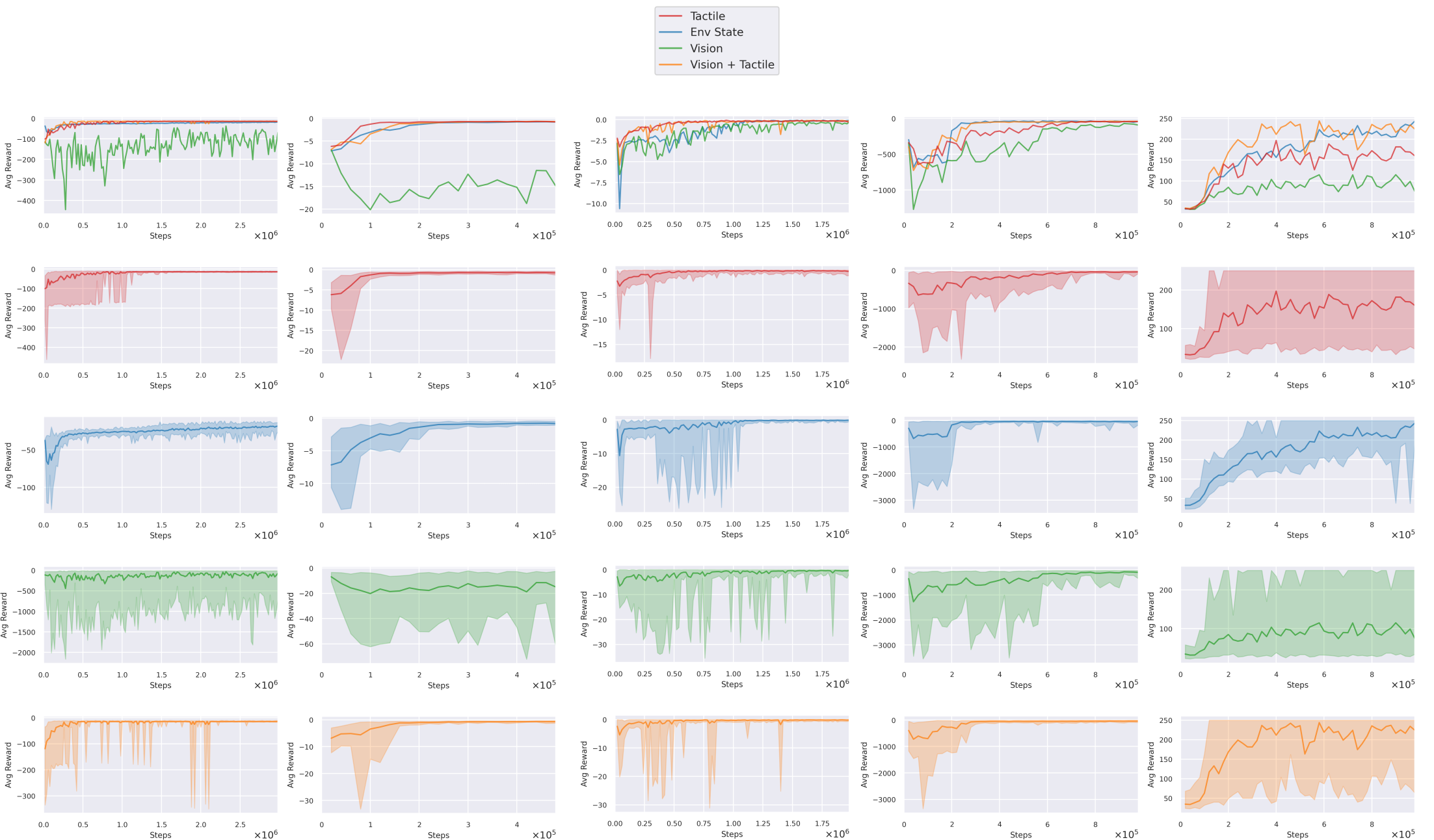}
          \put(2,  51){\parbox{25mm}{\centering (a) Edge Follow}}
          \put(22, 51){\parbox{25mm}{\centering (b) Surface Follow}}
          \put(42, 51){\parbox{25mm}{\centering (c) Object Roll}}
          \put(62, 51){\parbox{25mm}{\centering (d) Object Push}}
          \put(82, 51){\parbox{25mm}{\centering (e) Object Balance}}
      \end{overpic}
      \caption{Full evaluation of reinforcement learning agents throughout training. 10 evaluation episodes occur every 20,000 steps, using deterministic actions. Results averaged over 3 seeds. Shaded regions indicate maximum and minimum reward achieved over the 3 seeds.}
      \label{fig:full_rl_eval_results}
    \end{figure}
    
\newpage

\section{Pix2Pix Architecture} \label{apdx:pix2pix_arch}

\begin{table}[!ht]
    \scriptsize
    \setlength{\tabcolsep}{1pt}
    
    \begin{minipage}{.35\linewidth}

	\centering
	\caption{\label{tab:pix2pix_details} Pix2Pix architecture and parameters.}
	
	\begin{tabular}{|c|l|cccc|}
		\hline
		& \multicolumn{1}{l|}{ \textbf{Layer} } &  \multicolumn{4}{c|}{ \textbf{Details} }   \\ 
		\hline 
		
		\multirow{5}{*}{\STAB{\rotatebox[origin=c]{90}{ \textbf{Params}}}}
		& Batch Size    & \multicolumn{4}{c|}{64}           \\
		& Learning Rate & \multicolumn{4}{c|}{0.0002}    \\
		& Image Norm    & \multicolumn{4}{c|}{True}    \\
		& Image Trans   & \multicolumn{4}{c|}{[2.5\%, 2.5\%]}    \\
		& Loss Weights  & \multicolumn{4}{c|}{Wgan: 1.0, Wpix: 100.0]}    \\
		\hline
		
		\multirow{18}{*}{\STAB{\rotatebox[origin=c]{90}{ \textbf{Generator}}}}
		& Input dim        & \multicolumn{4}{c|}{ [128, 128, 1] }    \\
		& Output dim       & \multicolumn{4}{c|}{ [128, 128, 1] }    \\
		&  & \textbf{Input} & \textbf{Output} & \textbf{Dropout} & \textbf{Norm} \\
		& Down 1     &  1   & 64  & None & False \\
		& Down 2     &  64  & 128 & None & True \\
		& Down 3     &  128 & 256 & None & True \\
		& Down 4     &  256 & 512 & 0.5  & True \\
		& Down 5     &  512 & 512 & 0.5  & True \\
		& Down 6     &  512 & 512 & 0.5  & True \\
		& Down 7     &  512 & 512 & 0.5  & True \\
		& Up 1       &  512  & 512 & 0.5 &  True \\
		& Up 2       &  1024 & 512 & 0.5 &  True \\
		& Up 3       &  1024 & 512 & 0.5 &  True \\
		& Up 4       &  1024 & 256 & 0.5 &  True\\
		& Up 5       &  512  & 128 & None &  True \\
		& Up 6       &  256  & 64  & None &  True\\
		\hline 
		
		\multirow{7}{*}{\STAB{\rotatebox[origin=c]{90}{ \textbf{Discriminator}}}}
		& Input dim        & \multicolumn{4}{c|}{ [128, 128, 2] }    \\
		& Output dim       & \multicolumn{4}{c|}{ [16, 16, 1] }    \\
		&                  & \textbf{Input} & \textbf{Output} & \textbf{Norm} & \\
		& Disc 1           &  2    & 64   &   False & \\
		& Disc 2           &  64   & 128  &   True  & \\
		& Disc 3           &  128  & 256  &   True  & \\
		& Disc 4           &  256  & 512  &   True  & \\
		\hline
		
	\end{tabular}
	\end{minipage}\hfill
	\begin{minipage}{.6\linewidth}
        \centering
        \begin{overpic}[width=1.0\linewidth]{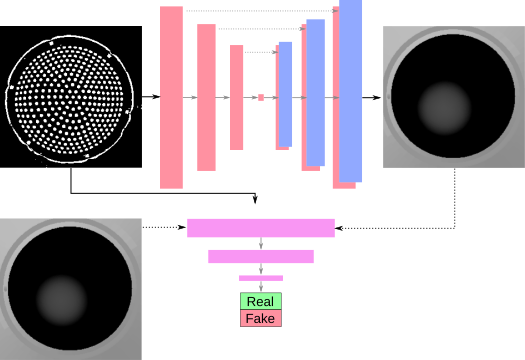}
            \put(-7, 65){\parbox{35mm}{\centering Real Image}}
            \put(-7,  28){\parbox{35mm}{\centering Target Image}}
            \put(66,   65){\parbox{35mm}{\centering Generated Image}}
            \put(43,  68){\parbox{35mm}{Generator}}
            \put(41.5, 27.75){\parbox{35mm}{Discriminator}}
        \end{overpic}
        \vspace{.25em}
        \captionof{figure}{\label{fig:pix2pix_arc}Real-to-sim translation of the tactile images uses a pix2pix-trained GAN. Real tactile images are processed by the generator to produce images that match the target simulated tactile images. The Discriminator is tasked with detecting whether an input tactile image pair is real or fake.}
	\end{minipage}

\end{table}

\section{Image Translation Data Collection} \label{apdx:pix2pix}

\begin{wrapfigure}{r}{0.475\linewidth}
    \centering
    \scriptsize
    \vspace{-1.35cm}
      \begin{overpic}[width=.95\linewidth]{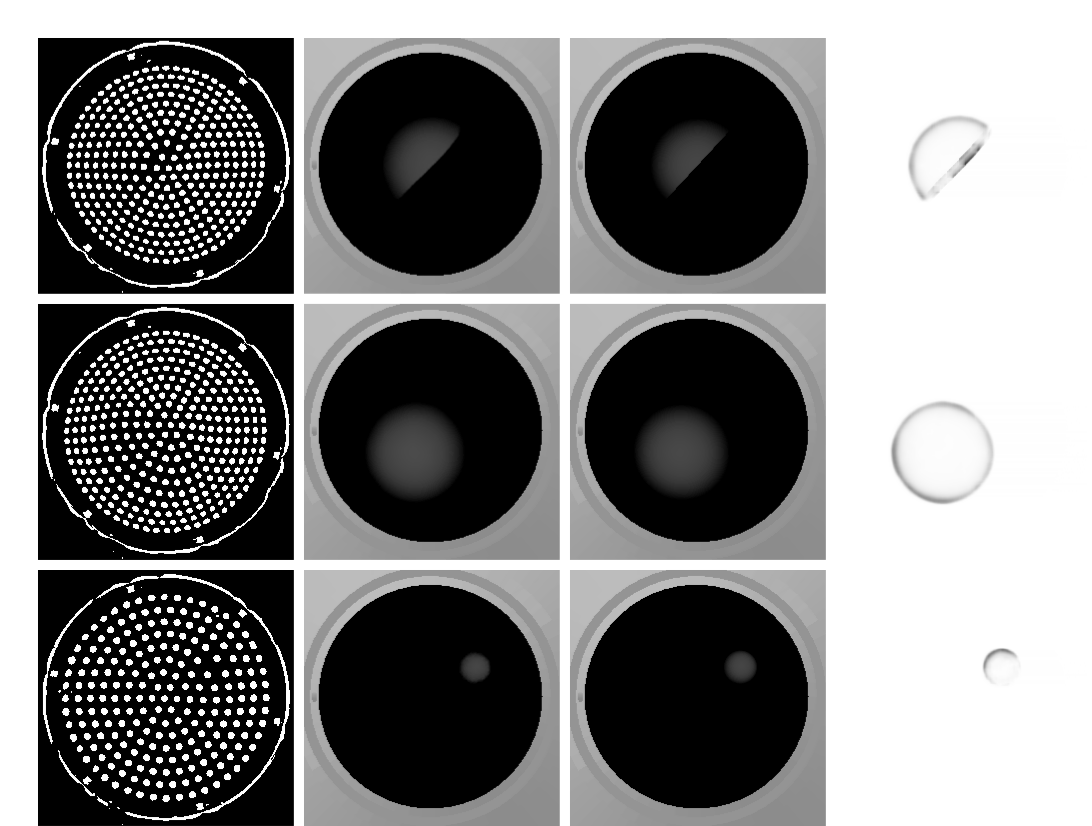}
          \put(-2,   73){\parbox{20mm}{\centering Real}}
          \put(23,  73){\parbox{20mm}{\centering Generated}}
          \put(48,  73){\parbox{20mm}{\centering Simulated}}
          \put(70,  73){\parbox{20mm}{\centering Difference}}
          
          \put(-1, 56){\rotatebox{90}{Edge}}
          \put(-1, 30){\rotatebox{90}{Surface}}
          \put(-1, 6 ){\rotatebox{90}{Probe}}
      \end{overpic}
  \caption{Image comparison between pairs of generated and simulated tactile images. Images are from validation sets for Real-to-Sim image translation. SSIM is used to create difference images.}
  \label{fig:gan_image_comparison}
\end{wrapfigure}
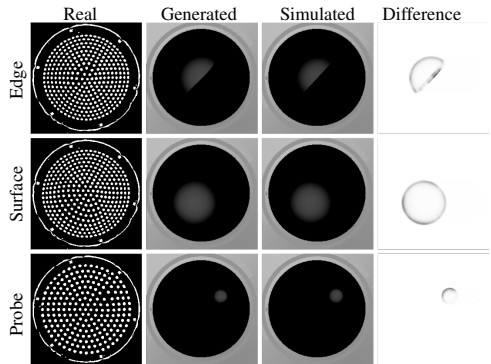

For the edge-following environment, we collect tactile images pressed onto a straight edge, varying the orientations, radial displacements and penetration of the sensor. A hemispherical sensor tip is used with the tool center point (TCP) located centrally at the end of the sensor. Relative to the TCP, the data is gathered over ranges: orientation $Rz \in [-179\si{\degree},180\si{\degree}]$, radial displacement $y \in [-6,6]\si{\mm}$, and penetration $z \in [3.5,5.5]\si{\mm}$.

For the surface-following and object-pushing environments, we collect tactile images pressed onto a flat surface, varying the orientations and penetration, also with a hemispherical tip. Relative to the TCP, data is gathered over ranges: orientation $\{Rx,Ry\} \in [-15\si{\degree},15\si{\degree}]$, and penetration $z \in [2,5]\si{\mm}$. 

For the object-rolling environment, we collect tactile images pressed onto a spherical probe stimulus, using a flat sensor tip appropriate to this environment. 9 spherical probe stimuli are used, ranging over $[2,6]\si{\mm}$ radius in $0.5\si{\mm}$ increments. The sensor is positioned to contact the probe at random placements within a $15\si{\mm}$ disk surrounding the centre of the tip. In this case, shear is not introduced into the data collection because rolling objects induce negligible motion-dependent shear.

\subsection{Full SSIM Scores} \label{apdx:ssim_scores}
We measure the SSIM scores across the validation sets collected for each task, each consisting of 2000 image pairs. For the Edge, Surface and Probe datasets respectively, we found mean scores of $[0.995, 0.992, 0.994]$, min scores of $[0.985, 0.982, 0.974]$, and max scores of $[0.999, 0.999, 0.999]$. Whilst high scores are expected due to the relatively sparse target images, this indicates strong performance in all cases.

\newpage
\section{Supervised Learning Comparison} \label{apdx:supervised_comparison}

\begin{wraptable}{r}{0.5\linewidth}
\footnotesize
\caption{\label{tab:sup_sim2real_comparison} Mean Absolute Error (MAE) for Task I: predicting polar angle (radians), Task II: predicting radial displacement (\si{\mm}) and Task III*: predicting position of a probe (\si{\mm}).}
\begin{center}
    	\begin{tabular}{|c|c|c|c|}
    		\hline
    		Approach & Task I & Task II & Task III*    \\ 
    		\hline 
    		Ours  & \textbf{0.079} & \textbf{0.119} & \textbf{0.059} \\
    		\citet{Ding2020Sim-to-RealSensing} & 0.254  & 0.45  & 0.73 \\
    		\hline
    	\end{tabular} \\
    \vspace{0.5em}	
    \scriptsize{*Indicates some difference between tasks as discussed below.}
\end{center}
\vspace{-1em}
\end{wraptable}

Whilst we were unable to find a sim-to-real reinforcement learning baseline that does not require specific hardware, we can compare to previous work simulating the TacTip sensor for supervised learning tasks. \citet{Ding2020Sim-to-RealSensing} used an elastic deformation approach to simulate the TacTip sensor. They focussed on supervised learning from simulated pin positions rather than tactile images, and could accurately predict edge position and orientation, and the $(x,y)$ position of a pole pressed into the sensor. In this section we draw a comparison between the presented method on their supervised learning tasks.

The tasks are defined as follows: Task I predicts polar angle $\theta$ of an edge pressed into the sensor; Task II predicts radial displacement $r$ of an edge pressed into the sensor; and Task III predicts the $(x,y)$ location of the centre of a pole pressed into the sensor. More details are provided in \cite{Ding2020Sim-to-RealSensing} . 

As this is not presented as a standardised benchmark, and we do not have access to the exact setup used, there are some differences in how these tasks have been carried out between the present work and \cite{Ding2020Sim-to-RealSensing}. In particular, we differ notably in Task III, where we use a flat TacTip sensor tip instead of the original hemispherical tip, because we are using the  data, GANs and objects acquired for the object rolling task which needed a flat tip. That said, we expect there will only be small differences in accuracy between flat and curved TacTip tips.

For each task we first train a convolutional neural network to predict $r$, $\theta$, $x$ and $y$ using a dataset of 10,000 simulated tactile images per task (taking $\sim$145 seconds to collect). We then collect a dataset of $2000$ real sensor images and using GANs trained on edge or probe data, we translate from real-to-sim images to create the test datasets of generated images. The performance of the trained networks over the full generated datasets is then measured and compared with the results from \cite{Ding2020Sim-to-RealSensing}.

As our simulated tactile data is comprised of images, we use a convolutional neural network (CNN). To avoid boundary issues over full rotations, we predict a sine/cosine encoding of angle. The CNN has 4 convolutional layers (each with kernel: 5, stride: 1, padding: 2, maxpool: 2) and 3 fully connected layers (dimensions = 1024, 512, output\_dim). Batch normalization is applied only to the convolutional layers before the ELU \cite{clevert2015fast} activation. Images of resolution $256\times256$-pixels were used.

Table \ref{tab:sup_sim2real_comparison} shows a large reduction in Mean Absolute Error (MAE) for both radial displacement and angle prediction in comparison with previous work simulating the TacTip \cite{Ding2020Sim-to-RealSensing}. A 3-fold improvement is found in predicting the angle $\theta$ (Task I) and a near 4-fold improvement when predicting radial distance (Task II). We also find a large decrease in MAE from 0.73\si{\mm} to 0.059\si{\mm} for position prediction (Task III), although as mentioned above there are some differences in setup.
    
\vfill

\end{document}